\pdfoutput=1

\documentclass[11pt]{article}

\usepackage[final]{EACL2023}

\usepackage{times}
\usepackage{latexsym}

\usepackage{graphicx}  
\usepackage{subfig}  
\usepackage{lipsum}  
\usepackage{booktabs}  
\usepackage{multirow}  
\usepackage{amsmath,amssymb,amsfonts,amsthm,bm,bbm}  

\usepackage[T1]{fontenc}

\usepackage[utf8]{inputenc}

\usepackage{microtype}

\usepackage{inconsolata}

\title{Few-Shot Structured Policy Learning \\ for Multi-Domain and Multi-Task Dialogues}

\author{Thibault Cordier\textsuperscript{1,2} \and Tanguy Urvoy\textsuperscript{2} \and Fabrice Lefèvre\textsuperscript{1} \and Lina M. Rojas-Barahona\textsuperscript{2} \\
  \textsuperscript{1}LIA - Avignon University, France \\
  \textsuperscript{2}Orange Innovation, Lannion, France \\
  \small\texttt{thibault.cordier@alumni.univ-avignon.fr} \\
  \small\texttt{fabrice.lefevre@univ-avignon.fr} \\
  \small\texttt{\{linamaria.rojasbarahona, tanguy.urvoy\}@orange.com} \\
}
\begin{document}
\maketitle
\begin{abstract}
\textit{Reinforcement learning} has been widely adopted to model \textit{dialogue managers} in task-oriented dialogues. However, the user simulator provided by state-of-the-art dialogue frameworks are only rough approximations of human behaviour.
The ability to learn from a small number of human interactions is hence crucial, especially on multi-domain and multi-task environments where the action space is large.
We therefore propose to use \textit{structured policies} to improve sample efficiency when learning on these kinds of environments.
We also evaluate the impact of \textit{learning from human vs simulated experts}.
Among the different levels of structure that we tested, the graph neural networks (GNNs) show a remarkable superiority by reaching a success rate above $80\%$ with only $50$ dialogues, when learning from simulated experts.
They also show superiority when learning from human experts, although a performance drop was observed,
indicating a possible difficulty in capturing the variability of human strategies.
We therefore suggest to concentrate future research efforts on bridging the gap between human data, simulators and automatic evaluators in dialogue frameworks.
\end{abstract}

\section{Introduction}
\label{sec:introduction}

Multi-domain multi-task dialogue systems are designed to complete specific \textit{tasks} in distinct \textit{domains} such as finding and booking
a hotel or a restaurant
\citep{zhu2020convlab}.
A domain is formally defined
as a list of \textit{slots} with their valid values. The most common task, the information-seeking task, is usually modelled as a slot-filling data-query problem in which the system requests constraints to the user and proposes items that fulfil those constraints.

The design of a \textit{dialogue manager} (\textsc{DM}s) is costly: \textit{hand-crafted} policies require a lot of engineering, pure \textit{supervised learning} (or \textit{behaviour cloning}) requires a lot of expert demonstrations, and pure \textit{reinforcement learning} requires a lot of user interactions to converge.
The simulators provided with
frameworks, such as \textsc{PyDial}~\cite{ultes2017pydial} or
\textsc{ConvLab}~\cite{zhu2020convlab},
are only rough approximations of human behaviour and
the ability to learn from a small number of human interactions remains crucial.
This is especially true on multi-domain and multi-task environments where the action space is large~\citep{gao2018neural}.

A popular approach to reduce these costs is to wire some knowledge about the problem into the policy model,
namely: \textit{few shot learning}
~\cite{wang2020generalizing}.
In particular, structured policies like \textit{graph neural networks} (\textsc{GNN}s) are known to be well suited to handle a variable number of slots and
domains for the information-seeking task
(\citealt{chen2018structured}; \citealt{chen2020distributed}).
In this paper, we explore structured policies based on \textsc{GNN}. A graph in a GNN is \textit{fully connected} and \textit{directed}. Each \textit{node} represents a sub-policy associated with a slot, while a directed \textit{edge} between two nodes represents a message passing.

For studying sample efficiency, we analyse the dialogue success rate of structured policies once trained in a supervised way from expert demonstrations.
We consider two types of demonstrations: \textit{human experts} extracted from the \textsc{MultiWOZ} dataset~\cite{budzianowski2018multiwoz}, and \textit{simulated experts} generated by letting the
\textsc{ConvLab}'s \textit{hand-crafted} policy interact with a simulated user.

We perform large scale experiments. We study the impact of different levels of structure (see them in Figure~\ref{fig:proposals}) on policy success rate after a limited number of dialogue demonstrations. For each level of structure, we also compare two sources of demonstrations: simulated and human dialogues.
We show a notable result: our structured policies are able to reach a success rate above $80\%$ with only $50$ when following a simulated expert in \textsc{ConvLab}. To the best of our knowledge there are not previous works that studied the impact of structure for dialogue policy in a few-shot setting.

Another important finding is that few-shot learning from human demonstrations is harder, producing a lower success rate.
This can be explained first by the large variability of human strategies that is not covered by simulated users which stick to more repetitive -- easy to learn -- dialogue patterns.
Another explanation could be an evaluation bias, simulated dialogues are more in line with artificial evaluators.

The remainder of this paper is structured as follows. We present the related work in Section~\ref{sec:related_work}. Section~\ref{sec:proposals} presents the proposed \textsc{GNN}s
from demonstrations.
The experiments and evaluation are described in Sections~\ref{sec:experiments} and~\ref{sec:evaluation} respectively. Finally, we conclude in Section~\ref{sec:conclusion}.

\section{Related Work}
\label{sec:related_work}

\textit{Few shot learning} takes advantage of prior knowledge to avoid overloading the empirical risk minimiser when the number of available examples is small. In particular, prior knowledge can be used to constrain hypothesis space (i.e. model parameters) with parameter sharing or tying in order to reduce reliance on data acquisition and on data annotation~\cite{wang2020generalizing}.

Prior knowledge can be built into dialogue systems by imposing a structure in the neural network architecture.
A first approach is to use \textit{hierarchical reinforcement learning} that divides a main problem into several simpler sub-problems.
We refer to \citet{sutton1999between} that introduces \textit{semi-Markov decision process} using temporal abstraction and to \citet{wen2020efficiency} that introduces \textit{sub-Markov decision process} using state partition. In the scope of the paper, a \textit{hierarchical policy} corresponds to a meta-controller that chooses to activate a domain and we have one sub-policy per domain \citep{budzianowski2017sub, casanueva2018feudal, le2018hierarchical}.

In the same vein, \textit{graph neural networks} (\textsc{GNN}s) have been explored in a wide range of domains because of their empirical success and their theoretical properties which explains its efficiency: the abilities of generalisation, stability and expressiveness
~\cite{garcia2018few}.
\textsc{GNN}s are suitable for applications where the data have a graph structure i.e where the graph outputs are supposed to be permutation-invariant or equivariant to the input features \citep{zhou2020graph, wu2020comprehensive}.

In single-domain dialogue environments, this architecture has been adapted to model the \textsc{DM} in \citet{chen2018structured} and \citet{chen2020distributed}.
They have shown that \textsc{GNN}s generalise between similar dialogue slots, manage a variable number of slots and transfer to different domains that perform similar tasks.
We thus adopt in this work the \textit{domain independent parametrisation} (\textsc{DIP}) \citep{wang2015learning}, which standardises the slots representation into a common feature space.

{In this work, as in~\citet{chen2018structured} and \citet{chen2020distributed}, we propose to improve multi-domain covering by learning a generic policy based on \textsc{GNN}. But unlike them, (i) we use a multi-domain multi-task setting, in which several domains and tasks can be evoked in a dialogue; (ii) the \textit{dialogue state tracker} (\textsc{DST}) output is not discarded when activating the domain; and (iii) we adapt the \textsc{GNN} structure to each domain by keeping the relevant nodes while sharing the edge's weights.}

\section{Structured Policies with Expert Demonstrations}
\label{sec:proposals}

In order to investigate the impact of structured policies with behaviour cloning in improving sample efficiency in multi-domain multi-task dialogue environments,
we introduce the dialogue state and action spaces for structured policies and we present the different policies and the experts' nature.

\begin{figure*}[!t]
    \centering
    \includegraphics[width=0.99\textwidth]{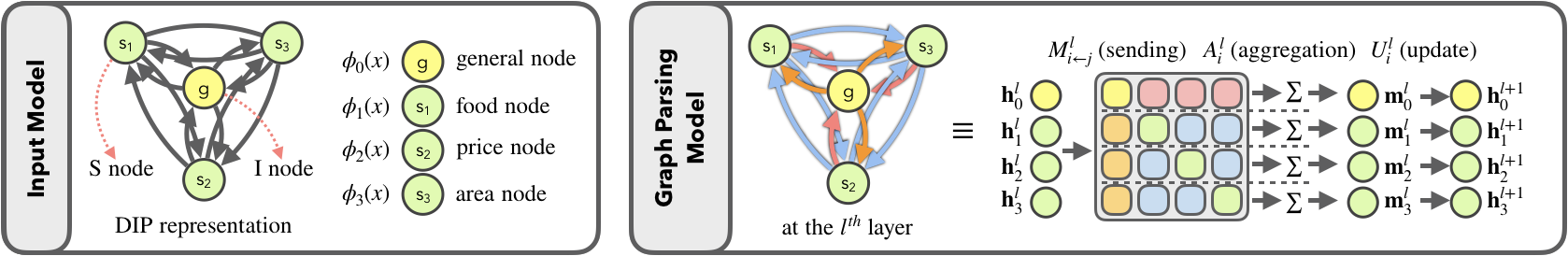}
    \caption{Structure of the input and graph parsing model in restaurant domain example. The input is a fully-connected graph with two kinds of nodes and three kinds of edges.
    The \textsc{I-node} are depicted in yellow;
    the \textsc{S-node} in green.
    The structured policy is described by successive graph convolutions
    composed of the shared weights $\mathbf{W}^l_{i,j}$.
    }
    \label{fig:models}
\end{figure*}

\subsection{Dialogue State / Action Representations}
\label{subsec:dip}

In multi-domain multi-task dialogues, the \textit{domain} refers to the set of concepts and values speakers can talk about. Examples of domains are restaurants, attractions, hotels, trains, etc.
A \textit{dialogue act} is a predicate that refers to the performative actions of speakers in conversations \cite{austin1975things}.
These actions are formalised as predicates like \textsc{inform} (\textit{i.e.}, affirm) or \textsc{request} with slots or slot-values pairs as arguments.
Examples of system actions are: \textsc{request}(food), or \textsc{inform}(address). These structured actions are used to frame a message to the user.
We adopt here the multi-task setting as presented in \textsc{ConvLab}~\cite{zhu2020convlab}, in which a single dialogue can have the following tasks: (i) \textit{find}, in which the system requests information in order to query a database and make an offer; (ii) \textit{book}, in which the system requests information in order to book the item.

We adopt the \textsc{DIP} state and action representations, which are not reduced to a flat vector but to a set of sub-vectors: one corresponding to the domain parametrisation (or \textit{slot-independent representation}),
the others to the slots parametrisation (or \textit{slot-dependent representations}).
For any active domain, the input to the \textit{slot-independent representation} is the concatenation of the previous \textit{slot-independent} user and system actions (see examples of the output below, and a formal definition in Section~\ref{subsec:graph_neural_network}), the number of entities fulfilling the user's constraints in the database, the booleans indicating if the dialogue is terminated and whether an offer has been found / booked. The output corresponds to action scores such as \textsc{reqmore}, \textsc{offer}, \textsc{book}, \textsc{great}, etc.
Regarding the \textit{slot-dependent representation}, its input is composed of the previous \textit{slot-dependent} user and system actions (see output below), the booleans indicating if a value is known and whether the slot is needed for the \textit{find} / \textit{book} tasks. Its output are actions scores such as \textsc{inform}, \textsc{request} and \textsc{select}.
The parameterisation used
in \textsc{ConvLab} does not depend on the probabilistic representation of the states, \textit{i.e.} does not consider the uncertainty in the predictions made by the \textit{natural language understanding} (\textsc{NLU}) module.

\subsection{Graph Neural Network}
\label{subsec:graph_neural_network}

\begin{figure*}[!ht]
    \begin{center}
        \subfloat[\scriptsize{\textsc{FNN}.}]{
        \includegraphics[width=0.24\textwidth]{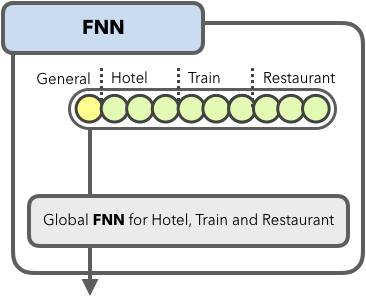}
        \label{subfig:FNN}
        }
        \subfloat[\scriptsize{\textsc{HFNN}.}]{
        \includegraphics[width=0.24\textwidth]{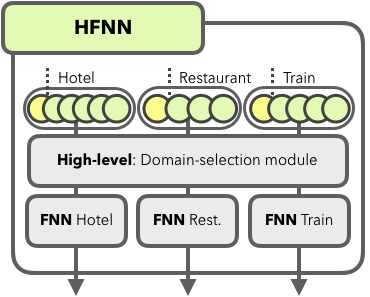}
        \label{subfig:HFNN}
        }
        \subfloat[\scriptsize{\textsc{HGNN}.}]{
        \includegraphics[width=0.24\textwidth]{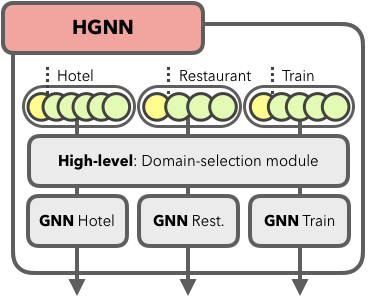}
        \label{subfig:HGNN}
        }
        \subfloat[\scriptsize{\textsc{UHGNN}.}]{
        \includegraphics[width=0.24\textwidth]{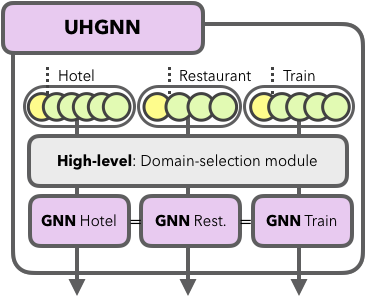}
        \label{subfig:UHGNN}
        }
    \end{center}
    \caption{Policy and input data structures. Different levels of structure are presented from classical \textit{feed-forward neural network} (\textsc{FNN}) to \textit{graph neural network} (\textsc{GNN}). The prefix \textsc{H-} corresponds to a hierarchical policy and \textsc{UH-} corresponds to a unique sub-policy for all domains. For a \textsc{FNN} layer, the input data is the concatenation of all \textsc{DIP} slot representations. For a \textsc{GNN} layer, the input keeps its structure.}
    \label{fig:proposals}
\end{figure*}

Prior knowledge can be integrated in our models by constraining the layer structure imposing symmetries in the neural dialogue policies. Without prior knowledge, the standard structure used is the \textit{feed-forward neural network} layer (\textsc{FNN}).
This unconstrained structure does not assume any symmetry in the network.

Assuming that sub-policies associated with the slots are the same, a better alternative is to use the \textit{graph neural network} layer (\textsc{GNN}).
This structure assumes that the state and action representations have a graph structure that are identically parameterised by \textsc{DIP}.
The \textsc{GNN} structure is a fully connected and directed graph, in which each \textit{node} represents a sub-policy associated with a slot and a directed \textit{edge} between two sub-policies represents a message passing.
We identify two roles for sub-policies: the general node as \textsc{I-node} associated to the \textit{slot-independent representation} and the slot nodes denoted as \textsc{S-node} associated to the \textit{slot-dependent representations}. Both representations were introduced in Section~\ref{subsec:dip}. We also identify the relations: \textsc{I2S} for \textsc{I-node} to \textsc{S-node}, \textsc{S2I} and \textsc{S2S} respectively\footnote{We omit the I2I relation because there is only one I-node.} (as presented in Figure~\ref{fig:models}).

We formally define the GNN structure as follows. Let $n$ be the number of slots and $L$ the number of layers. Let be $x$ the dialogue state, $\mathbf{x}_0=\phi_0(x)$, $\mathbf{h}^l_0\; \forall l\in[0,L-1]$ and $\mathbf{y}_0$ be respectively the input, hidden and output \textsc{I-node} representations.
Let the input, hidden and output \textsc{S-nodes} representations be respectively $\forall i \in[1,n], \, \mathbf{x}_i=\phi_i(x)$, $\mathbf{h}^l_i\ \forall l\in[0,L-1]$ and $\mathbf{y}_i$.
First, the \textsc{GNN} transforms inputs:
\begin{eqnarray}
    \textstyle \forall i \in [0, n],\quad \mathbf{h}^0_i = \sigma^0(\mathbf{W}_i^0 \phi_i(\mathbf{x})+ \mathbf{b}_i^0)
\end{eqnarray}
Then, at the $l$-th layer, it computes the hidden nodes representations
by following message sending\footnote{The notation $i \leftarrow j$ denotes a message sending from slot $j$ to slot $i$. It also corresponds to the directed relation between the slots $j$ and $i$. The notation $i \leftarrow *$ denotes all messages sending to slot $i$.} (Eq.~\ref{eq:send}), message aggregation (Eq.~\ref{eq:agg}) and representation update (Eq.~\ref{eq:update}). $\forall i,j \in [0, n]^2$:
\begin{eqnarray}
    &\textstyle \mathbf{m}^{l}_{i \leftarrow j} = M_{i \leftarrow j}^l(\mathbf{h}^{l-1}_{j}) = \mathbf{W}_{i,j}^l \mathbf{h}^{l-1}_{j} + \mathbf{b}^{l}_{i,j}
    \label{eq:send} \\
    &\textstyle \mathbf{m}^{l}_{i} = A_i^l(\mathbf{m}^{l}_{i \leftarrow *}) = \frac{1}{n} \sum_{j=0}^n \mathbf{m}^{l}_{i \leftarrow j} \label{eq:agg} \\
    &\textstyle \mathbf{h}^{l}_{i} = U_i^{l}(\mathbf{m}^{l}_{i}) =  \sigma^{l}(\mathbf{m}^{l}_{i}  \label{eq:update})
\end{eqnarray}
The message sending function $M_{i \leftarrow j}^l$ is a linear transformation with bias.
The message aggregation function $A_i^l$ is the average pooling function.
The representation update function $U_i^{l}$ compute the new hidden representation with \textsc{ReLU} activation function and dropout technique during learning stage.
Finally, the \textsc{GNN} concatenates ($\oplus$ symbol) all final nodes representations and computes the policy function with the Softmax activation function.
\begin{eqnarray}
    \textstyle \mathbf{y} = \sigma^L(\bigoplus_{i=0}^n \mathbf{W}_{i}^L \mathbf{h}_i^{L-1} + \mathbf{b}_i^{L})
\end{eqnarray}

\subsection{Structured Policies}
\label{subsec:structured_policies}

We propose a wide range of dialogue policies to study the impact of the structure in sample efficiency. An ablation study progressively adds some notion of hierarchy to \textsc{FNN}s to approximate the structure of \textsc{GNN}s. Similarly, we analyse the advantage of sharing a generic \textsc{GNN} among several domains versus specialising a \textsc{GNN} to each domain. Therefore, we propose from the least to the most constrained:
\begin{itemize}
    \item \textbf{Feed-forward Neural Network} (\textsc{\textbf{FNN}}) that is a classical feed-forward neural network with \textsc{DIP} parametrisation (Figure~\ref{subfig:FNN}).
    \item \textbf{Hierarchy of Feed-forward Neural Networks} (\textsc{\textbf{HFNN}}) that is a hierarchical policy with  hand-crafted domain-selection  and \textsc{FNN}s for each domain. Each domain has one corresponding \textsc{FNN} model (Figure~\ref{subfig:HFNN}).
    \item \textbf{Hierarchy of Graph Neural Networks} (\textsc{\textbf{HGNN}}) that is a hierarchical policy with hand-crafted
    domain-selection and \textsc{GNN}s. Each domain has one corresponding \textsc{GNN} model (Figure~\ref{subfig:HGNN}).
    \item \textbf{Hierarchy with Unique Graph Neural Network} (\textsc{\textbf{UHGNN}}) that is a \textsc{HGNN} with a unique \textsc{GNN} for all domains. Each domain shares the same \textsc{GNN} model (Figure~\ref{subfig:UHGNN}).
\end{itemize}

\subsection{The Expert's Nature}
\label{subsec:data}

Since our goal is to learn on observed demonstrations delivered by an expert, we propose to focus on policies that learn from both simulated and  human experts. For this purpose, we use the dataset \textsc{MultiWOZ}~\cite{budzianowski2018multiwoz} to follow human experts and the hand-crafted policy of \textsc{ConvLab}~\cite{zhu2020convlab} as the simulated expert.

\paragraph{Human expert} The \textsc{MultiWOZ} dataset is a large annotated and open-sourced collection of human-human chats that covers multiple domains and tasks. Nearly 10k dialogues have been collected by a \textit{Wizard-of-Oz} set-up at relatively low cost and with a small
time effort. However, different versions of this dataset corrected and improved the annotations \cite{eric2019multiwoz, zang-etal-2020-multiwoz, han2021multiwoz, ye2021multiwoz}. In this work, we use the \textsc{MultiWOZ} dataset integrated in \textsc{ConvLab} with extended user dialogue act annotations.

\begin{figure*}[ht!]
    \begin{center}
        \subfloat[\scriptsize{\textsc{FNN}.}]{
        \includegraphics[width=0.24\textwidth]{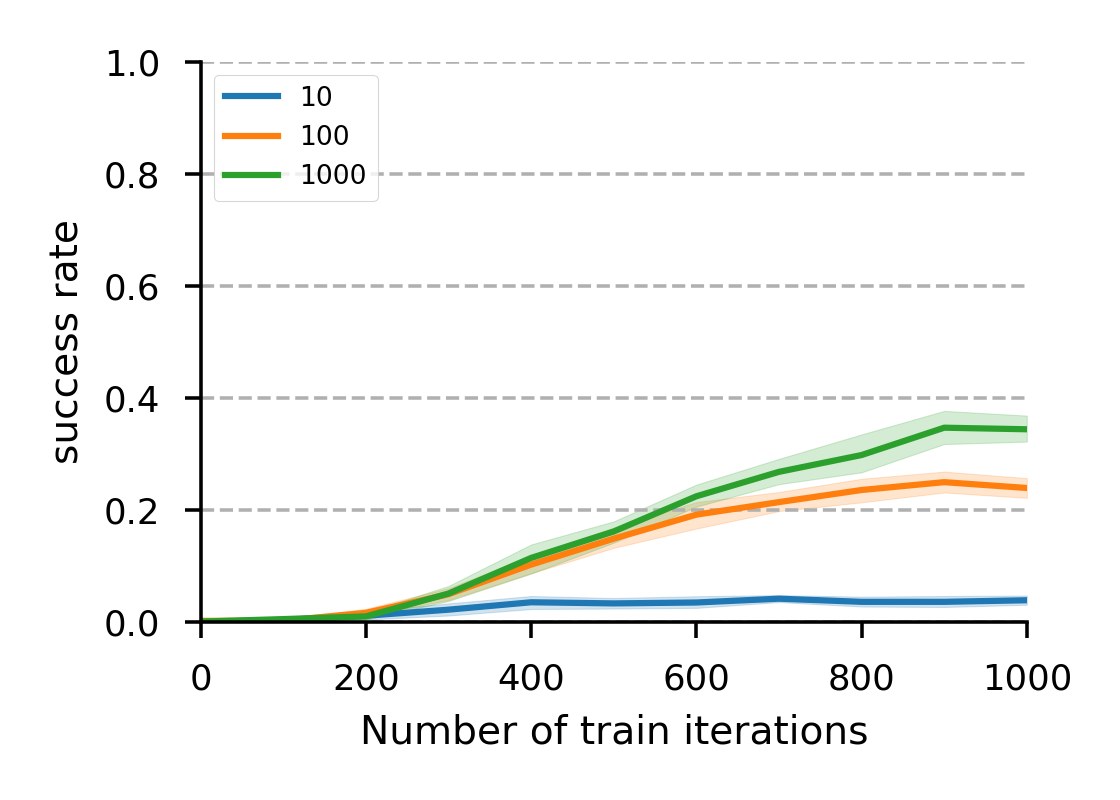}
        \label{subfig:success_FNN_CL_short}
        }
        \subfloat[\scriptsize{\textsc{HFNN}.}]{
        \includegraphics[width=0.24\textwidth]{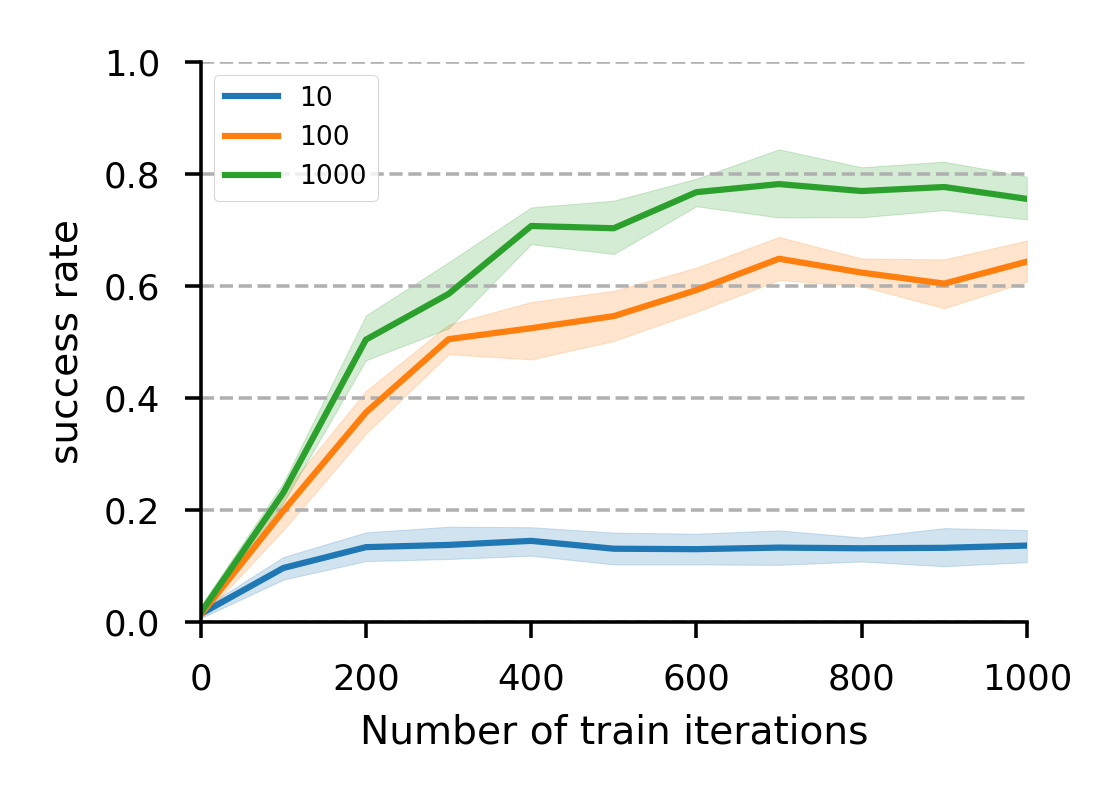}
        \label{subfig:success_HFNN_CL_short}
        }
        \subfloat[\scriptsize{\textsc{HGNN}.}]{
        \includegraphics[width=0.24\textwidth]{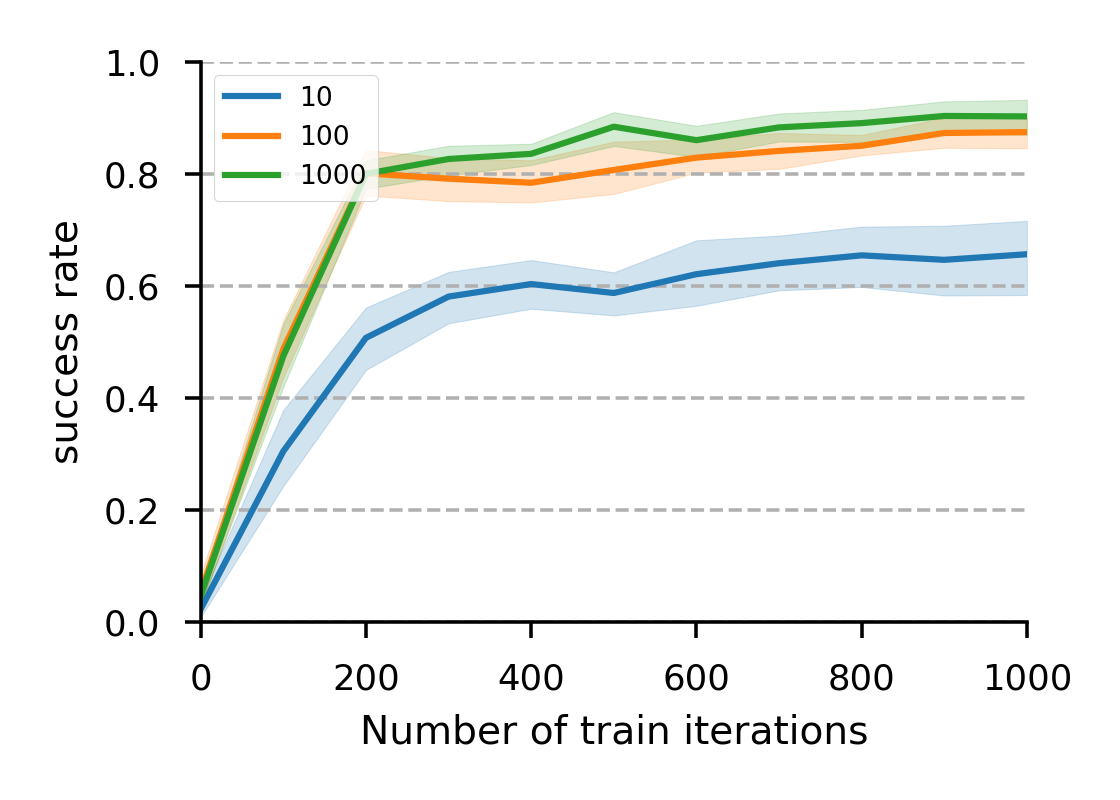}
        \label{subfig:success_HGNN_CL_short}
        }
        \subfloat[\scriptsize{\textsc{UHGNN}.}]{
        \includegraphics[width=0.24\textwidth]{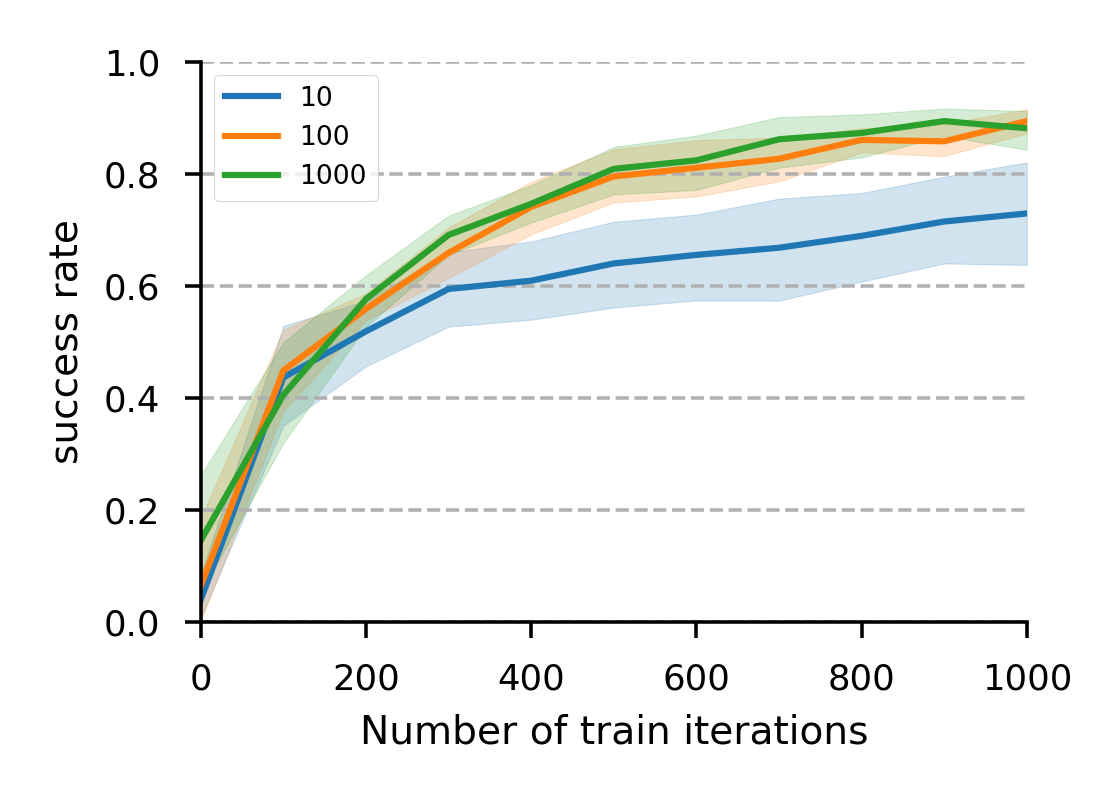}
        \label{subfig:success_UHGNN_CL_short}
        }
        \\
        \subfloat[\scriptsize{\textsc{FNN}.}]{
        \includegraphics[width=0.24\textwidth]{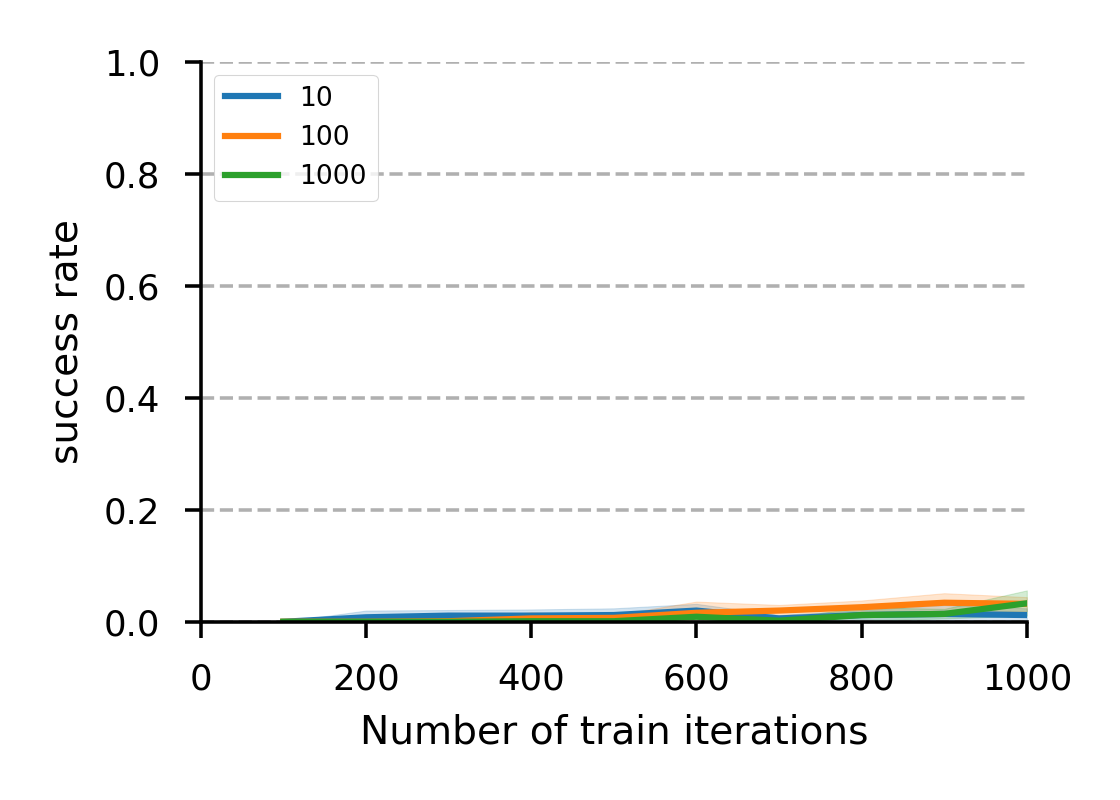}
        \label{subfig:success_FNN_MW_short}
        }
        \subfloat[\scriptsize{\textsc{HFNN}.}]{
        \includegraphics[width=0.24\textwidth]{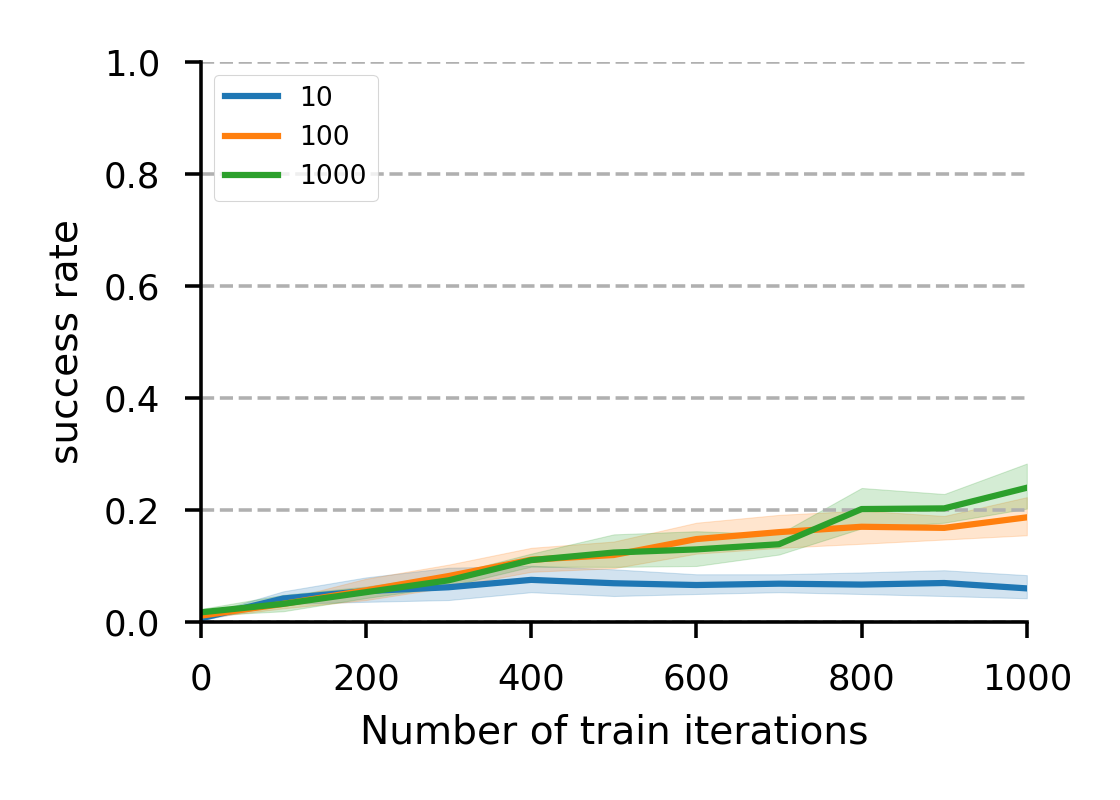}
        \label{subfig:success_HFNN_MW_short}
        }
        \subfloat[\scriptsize{\textsc{HGNN}.}]{
        \includegraphics[width=0.24\textwidth]{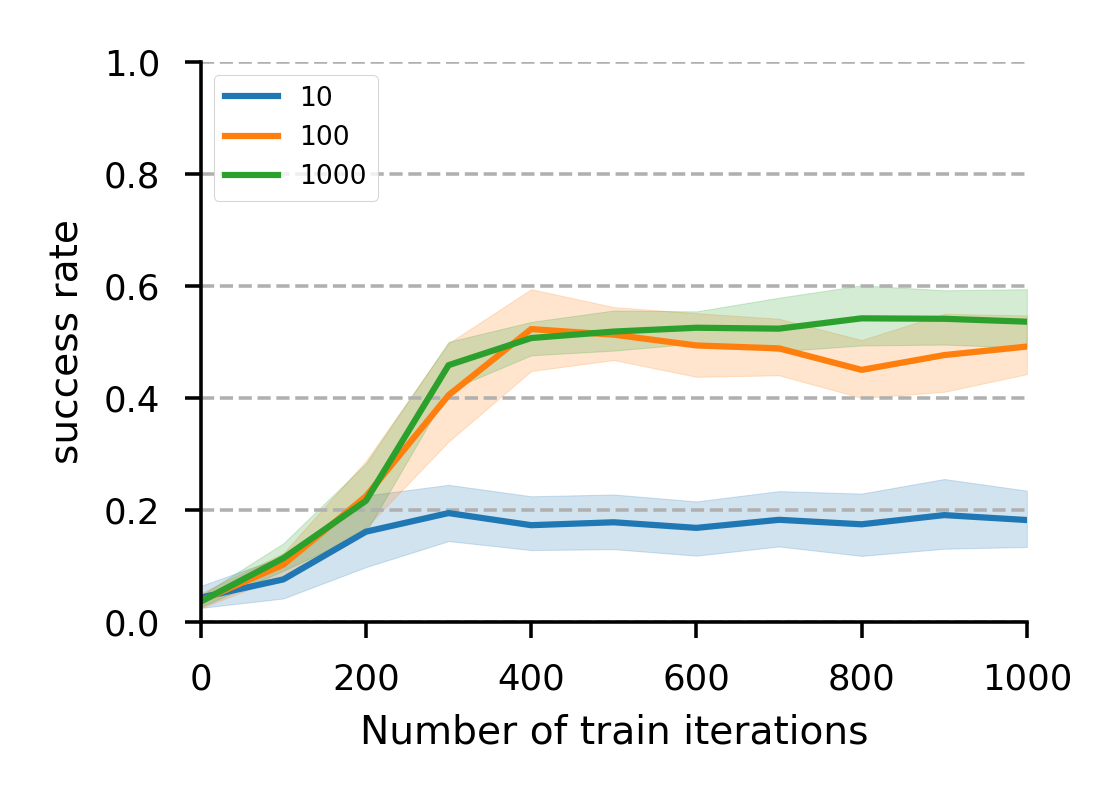}
        \label{subfig:success_HGNN_MW_short}
        }
        \subfloat[\scriptsize{\textsc{UHGNN}.}]{
        \includegraphics[width=0.24\textwidth]{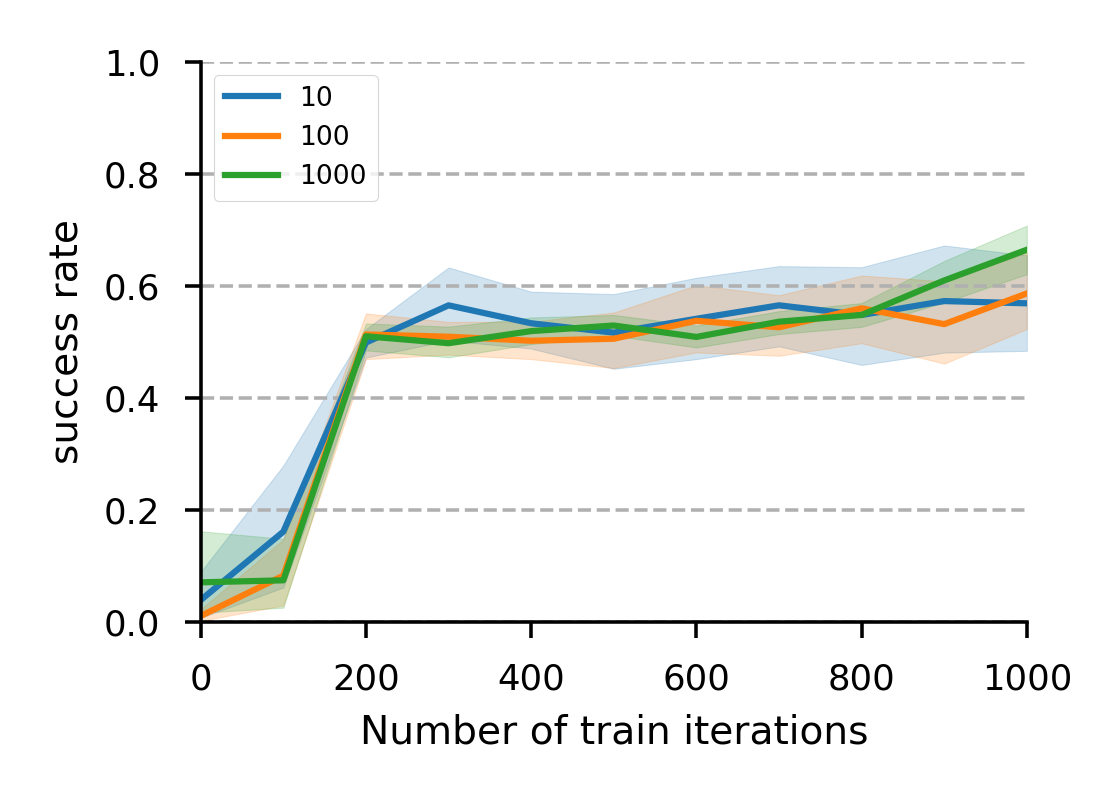}
        \label{subfig:success_UHGNN_MW_short}
        }
    \end{center}
    \caption{Dialogue manager evaluation with simulated users. We present the success rate on $10$ / $100$ / $1\,000$ training dialogues as a function of the number of gradient descent steps in a short training scenario. Learning is based on simulated experts (Figures (a) up to (d)) or on human experts (Figures (e) up to (h)). The line plot represents the mean and
    the coloured area represents the $95\%$ confidence interval over a sample of $10$ runs.}
    \label{fig:performance_short}
\end{figure*}

\paragraph{Simulated expert} The \textsc{ConvLab} framework has been proposed to automatically build, train and evaluate multi-domain multi-task oriented dialogue systems based on \textsc{MultiWOZ} features. It implements both hand-crafted simulated user and policy. The latter has been shown to be nearly the optimal policy according to the \textsc{ConvLab} evaluation setup of~\cite{takanobu2020your}. Therefore we use it as the simulated expert.

\section{Experiments}
\label{sec:experiments}

In this section we explain the experimental setup, the proposed models and the evaluation metrics.

\subsection{Experiment Setup}
\label{subsec:setup}

We performed an ablation study by gradually adding different levels of structure from a baseline \textsc{FNN} to the proposed \textsc{GNN} (Subsection~\ref{subsec:models}).
On the one hand, we analyse the learning efficiency of our models in small training steps. On the other hand, we compare their generalisation ability in few shot learning.

To analyse the learning efficiency, we measure performance
with respect to the number of gradient descent steps up to $1\,000$ iterations with a step size of $100$ iterations. We compare learning curves based on randomly chosen $10$, $100$ and $1\,000$ training dialogues\footnote{These values were chosen arbitrarily to give us an insight into the impact of the number of dialogues on the performance.}.
We also measure performance as a function of the number of training dialogues available (randomly chosen) namely $10$, $50$, $100$, $500$ and $1000$ when each training is performed up to $10\,000$ gradient descent steps.
All the experiments were run on \textsc{ConvLab}, restarted $10$ times with random initialisation and the results estimated on $500$ new dialogues.

\subsection{Models}
\label{subsec:models}

The \textsc{FNN} models have two hidden layers, both with $128$ neurons. The \textsc{GNN} models have one first hidden layer with $64$ neurons for both nodes (\textsc{S-node} and \textsc{I-node}). Then the second hidden layer is composed of $64$ neurons for each relation (\textsc{S2S}, \textsc{S2I} and \textsc{I2S}). For training stage, we use the \textsc{Adam} optimiser with a learning rate $lr = 0.001$, a dropout rate $dr = 0.1$ and a batch size $bs = 64$.

\begin{figure*}[ht!]
    \begin{center}
        \subfloat[\scriptsize{Success rate.}]{
        \includegraphics[width=0.32\textwidth]{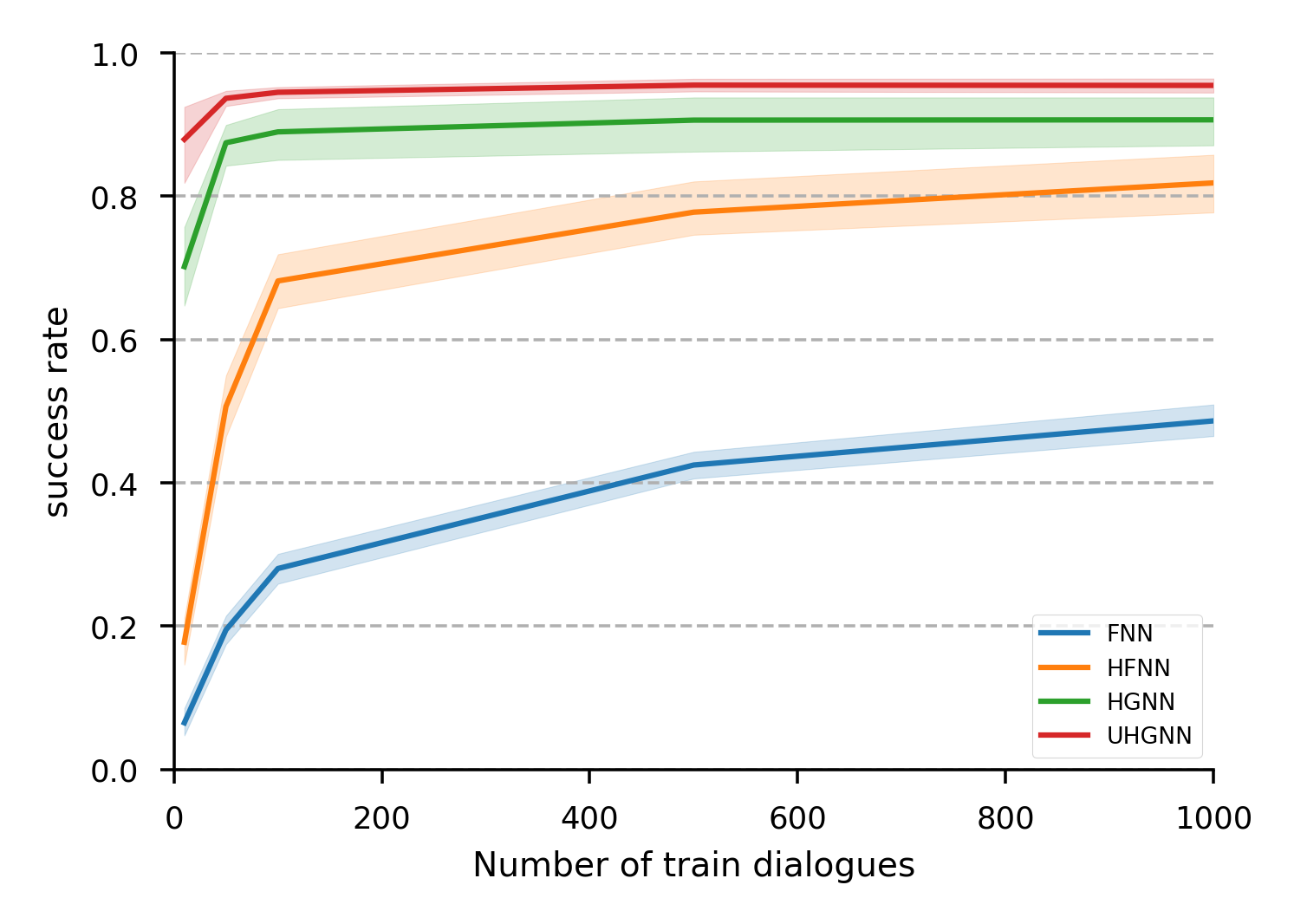}
        \label{subfig:success_CL_long}
        }
        \subfloat[\scriptsize{Inform rate (recall).}]{
        \includegraphics[width=0.32\textwidth]{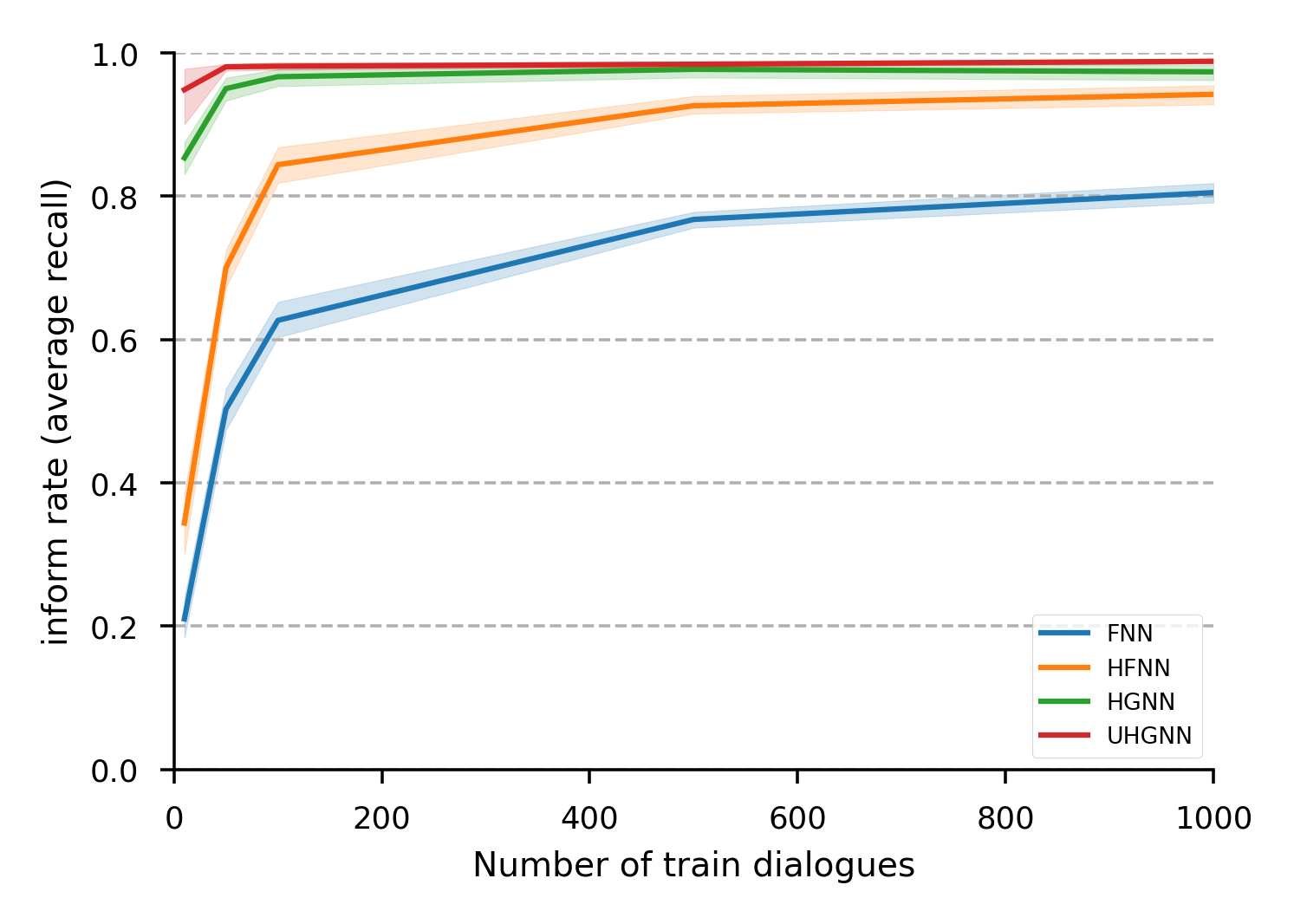}
        \label{subfig:find_CL_long}
        }
        \subfloat[\scriptsize{Book rate.}]{
        \includegraphics[width=0.32\textwidth]{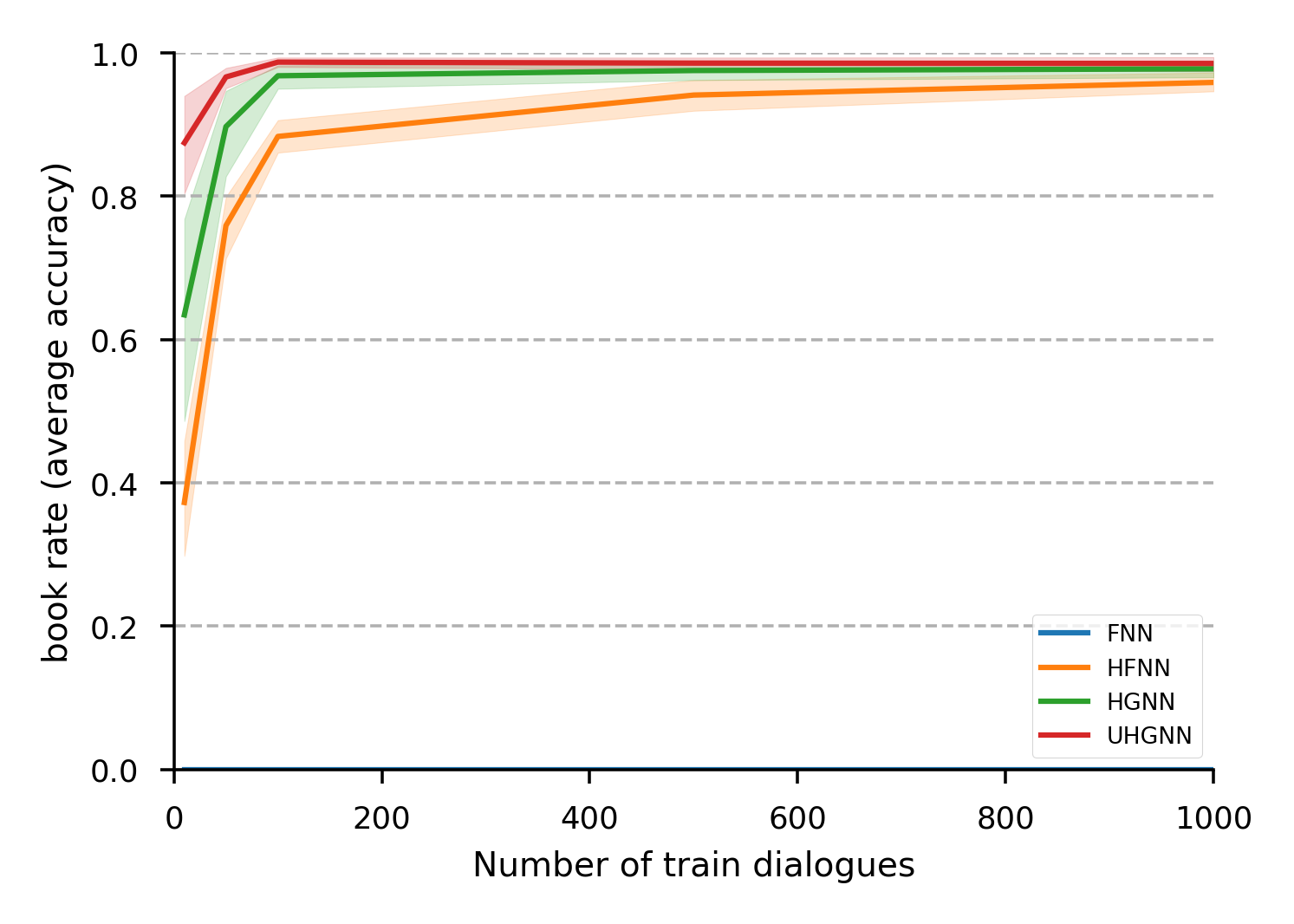}
        \label{subfig:book_CL_long}
        }\\
        \subfloat[\scriptsize{Success rate.}]{
        \includegraphics[width=0.32\textwidth]{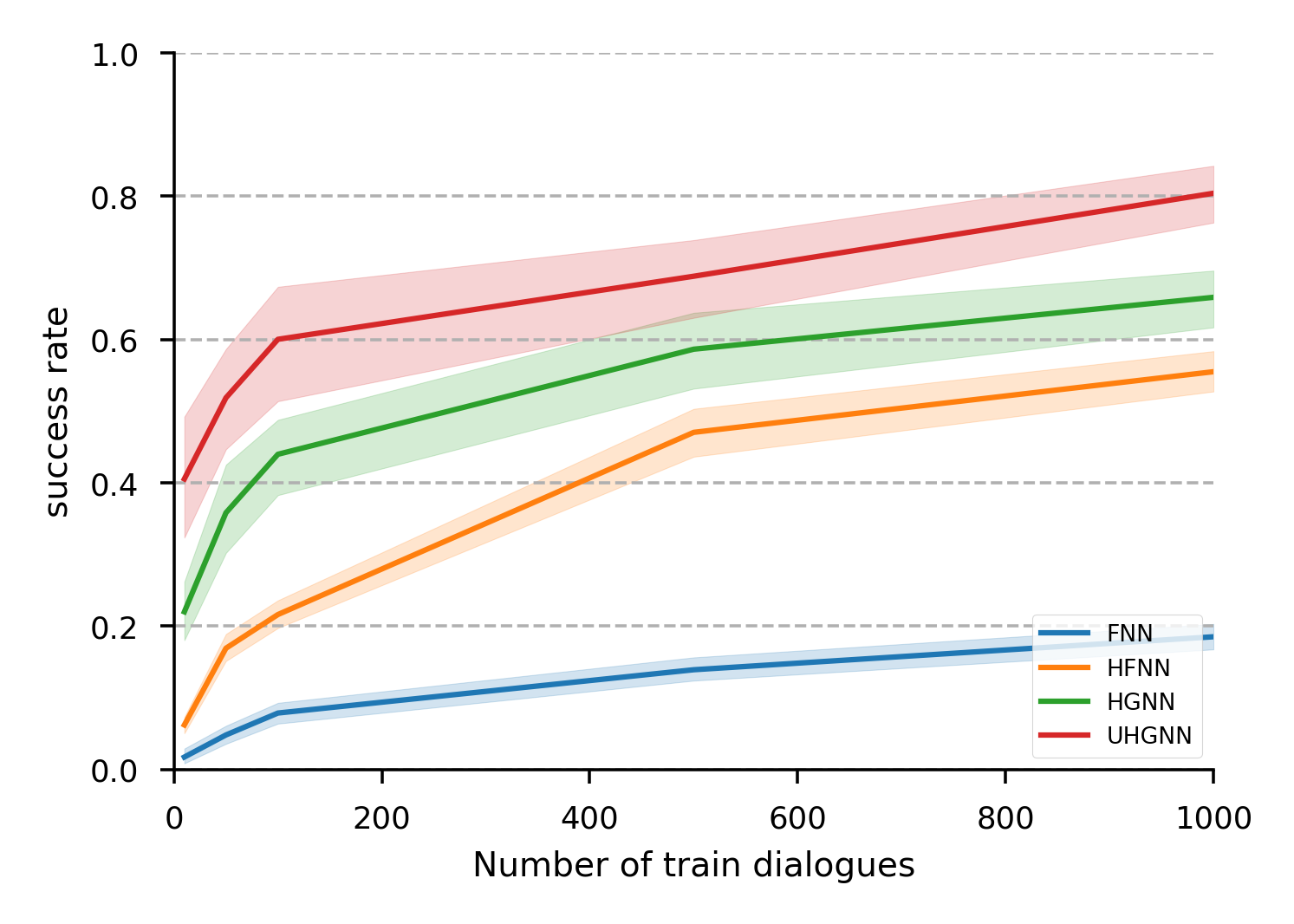}
        \label{subfig:success_MW_long}
        }
        \subfloat[\scriptsize{Inform rate (recall).}]{
        \includegraphics[width=0.32\textwidth]{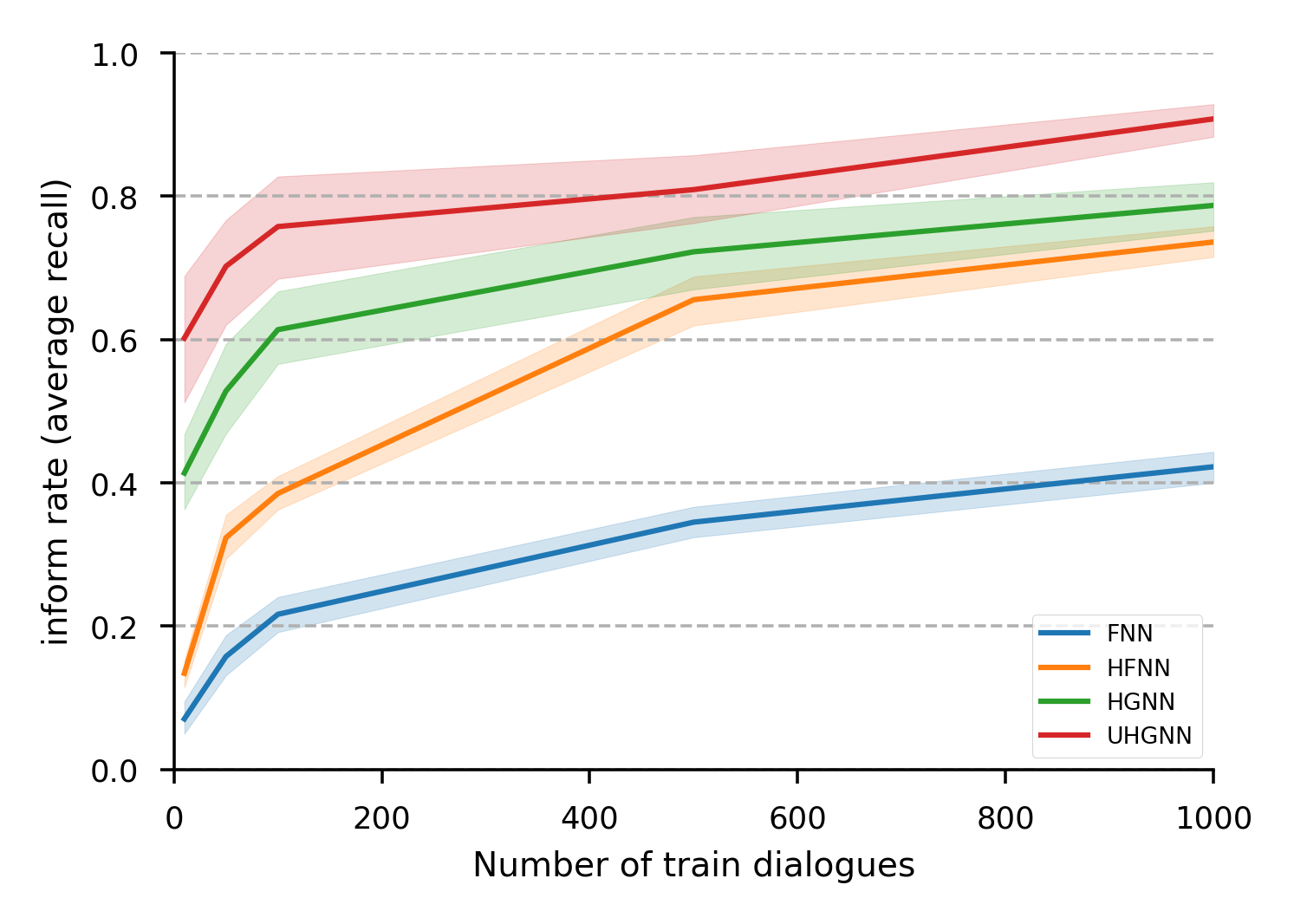}
        \label{subfig:find_MW_long}
        }
        \subfloat[\scriptsize{Book rate.}]{
        \includegraphics[width=0.32\textwidth]{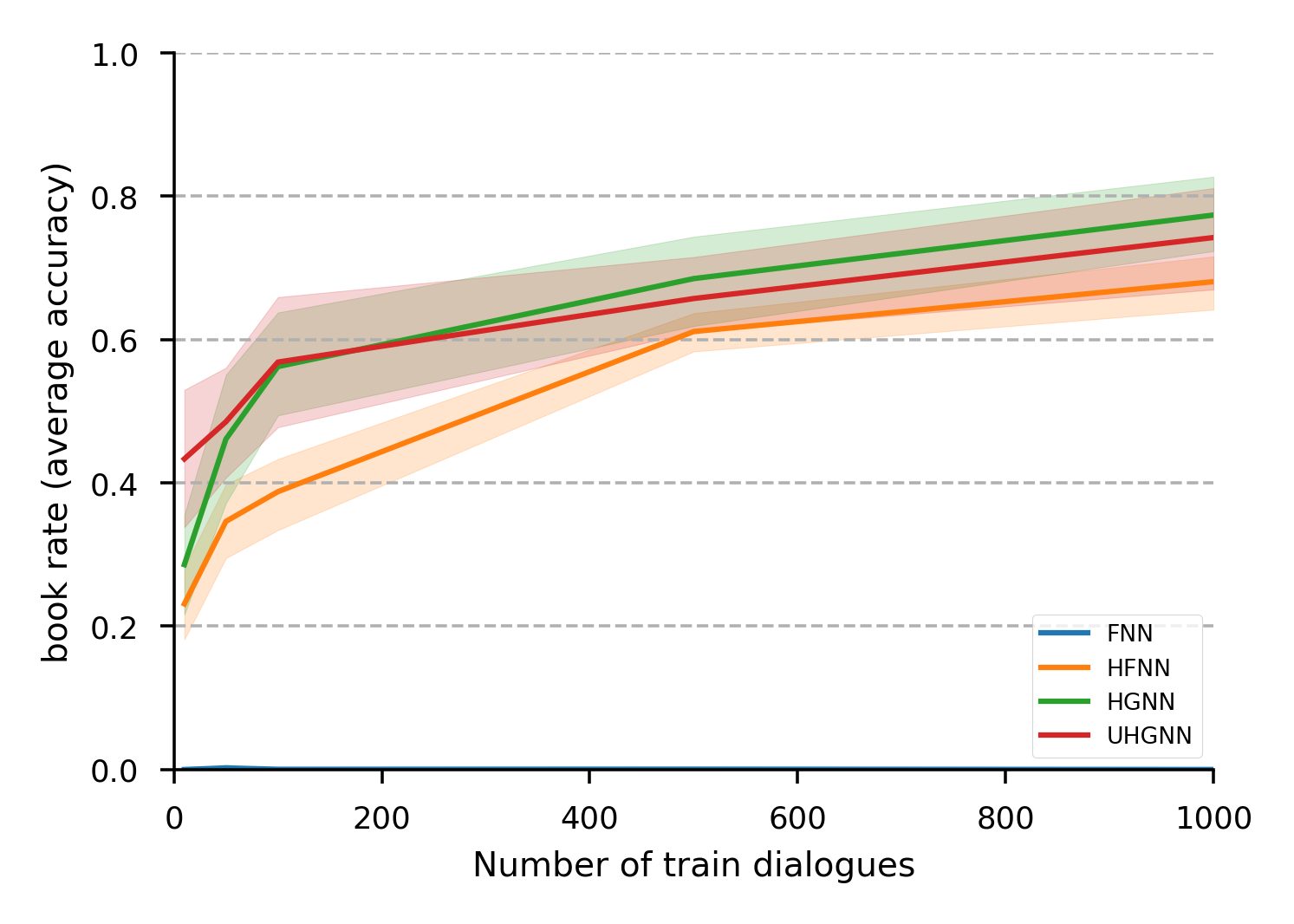}
        \label{subfig:book_MW_long}
        }
    \end{center}
    \caption{Dialogue manager evaluation with simulated user presenting the success rate based on $10\,000$ training iterations as a function of the number of training dialogues in a long learning scenario. Learning is based on a simulated expert (Figures (a), (b) and (c) ) or human experts (Figures (d), (e) and (f)). The line plot represents the mean and the coloured area represents the $95\%$ confidence interval over a sample of $10$ runs.}
    \label{fig:performance_long}
\end{figure*}

\subsection{Metrics}
\label{subsec:metrics}

We evaluate the performance of the policies for all tasks as in \textsc{ConvLab}.  Precision, recall and F-score, namely the \textbf{inform rates}, are used for the \textit{find} task. Inform recall evaluates whether all the requested information has been informed while inform precision evaluates whether only the requested information has been informed.
For the \textit{book} task, the accuracy, namely the \textbf{book rate}, is used.
It assesses whether the offered entity meets all the constraints specified in the user goal.
The dialogue is marked as \textbf{successful} if and only if both inform recall and book rate are equal to $1$.
The dialogue is considered \textbf{completed} if it is successful from the user's point of view\footnote{A dialogue can be completed without being successful if the information provided is not the one objectively expected by the simulator.}.

\section{Evaluation}
\label{sec:evaluation}

First, we evaluate the dialogue manager performance when talking to a simulated user. Second, we evaluate the learned policies within the entire dialogue system both with simulated and with real users. The evaluations have been done within \textsc{ConvLab}.

\subsection{Dialogue Manager Evaluation}
\label{subsec:dialogue_management_evaluation}

We analyse our models on the learning efficiency in small training steps and on the ability to generalise in a few-shot setting.

\paragraph{Efficiency}

We report in Figure~\ref{fig:performance_short}
the results of the ablation study
showing the ability of the models to succeed in a short training stage.
First, when learning from simulated demonstrations we notice in Figure~\ref{subfig:success_FNN_CL_short} that the baseline (\textsc{FNN}) needs a large number of training dialogues (more than $100$) to achieve a moderate performance (less than $40\%$). We show then in Figure~\ref{subfig:success_HFNN_CL_short} that hierarchical networks (\textsc{HFNN}) do improve learning efficiency up to $60\%$ with $100$ dialogues, up to $80\%$ with $1\,000$ dialogues.
Finally we show that graph neural network (\textsc{HGNN} in Figure~\ref{subfig:success_HGNN_CL_short}) and generic policy (\textsc{UHGNN} in Figure~\ref{subfig:success_UHGNN_CL_short}) drastically improve the efficiency with few dialogues, more than $60\%$ with $10$ dialogues, and achieve remarkable performance above $80\%$ with only $100$ dialogues in $1\,000$ training steps. These observations confirm that hierarchical and generic \textsc{GNN}s allow efficient learning and collaborative gradient update in a short training stage.

Although standard or hierarchical policies (\textsc{FNN} in Figure~\ref{subfig:success_FNN_MW_short} and \textsc{HFNN} in Figure~\ref{subfig:success_HFNN_MW_short}) are less efficient when learning from human demonstrations, they are still above baselines. It is worth noting that structured or generic \textsc{GNN} policies \textsc{HGNN} in Figure~\ref{subfig:success_HGNN_MW_short} and \textsc{UHGNN} in Figure~\ref{subfig:success_UHGNN_MW_short} are able to reach more than $50\%$ success rate.

\begin{figure*}[ht!]
    \begin{center}
        \subfloat[\scriptsize{Success rate.}]{
        \includegraphics[width=0.32\textwidth]{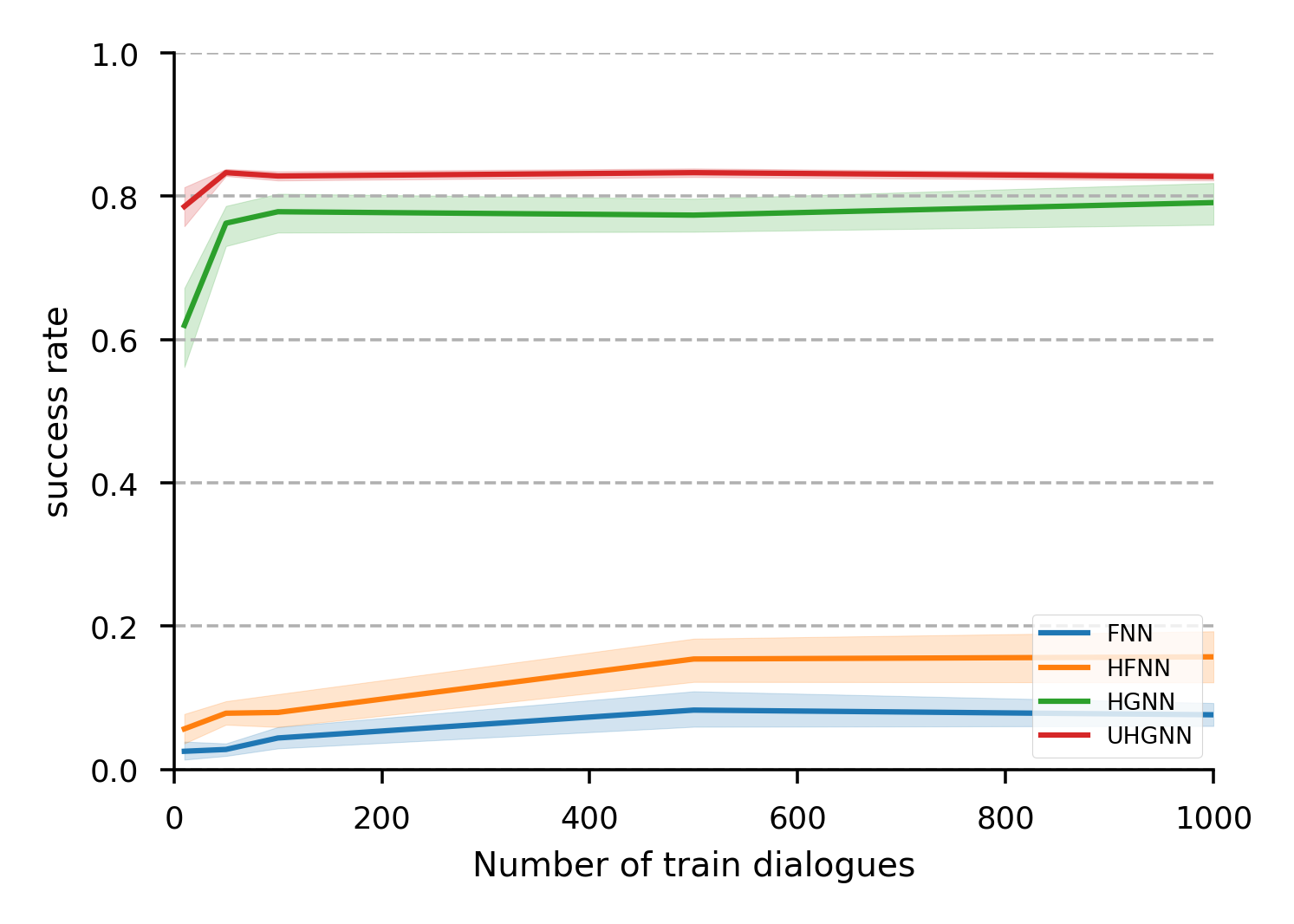}
        \label{subfig:success_CL_pipeline}
        }
        \subfloat[\scriptsize{Inform rate (recall).}]{
        \includegraphics[width=0.32\textwidth]{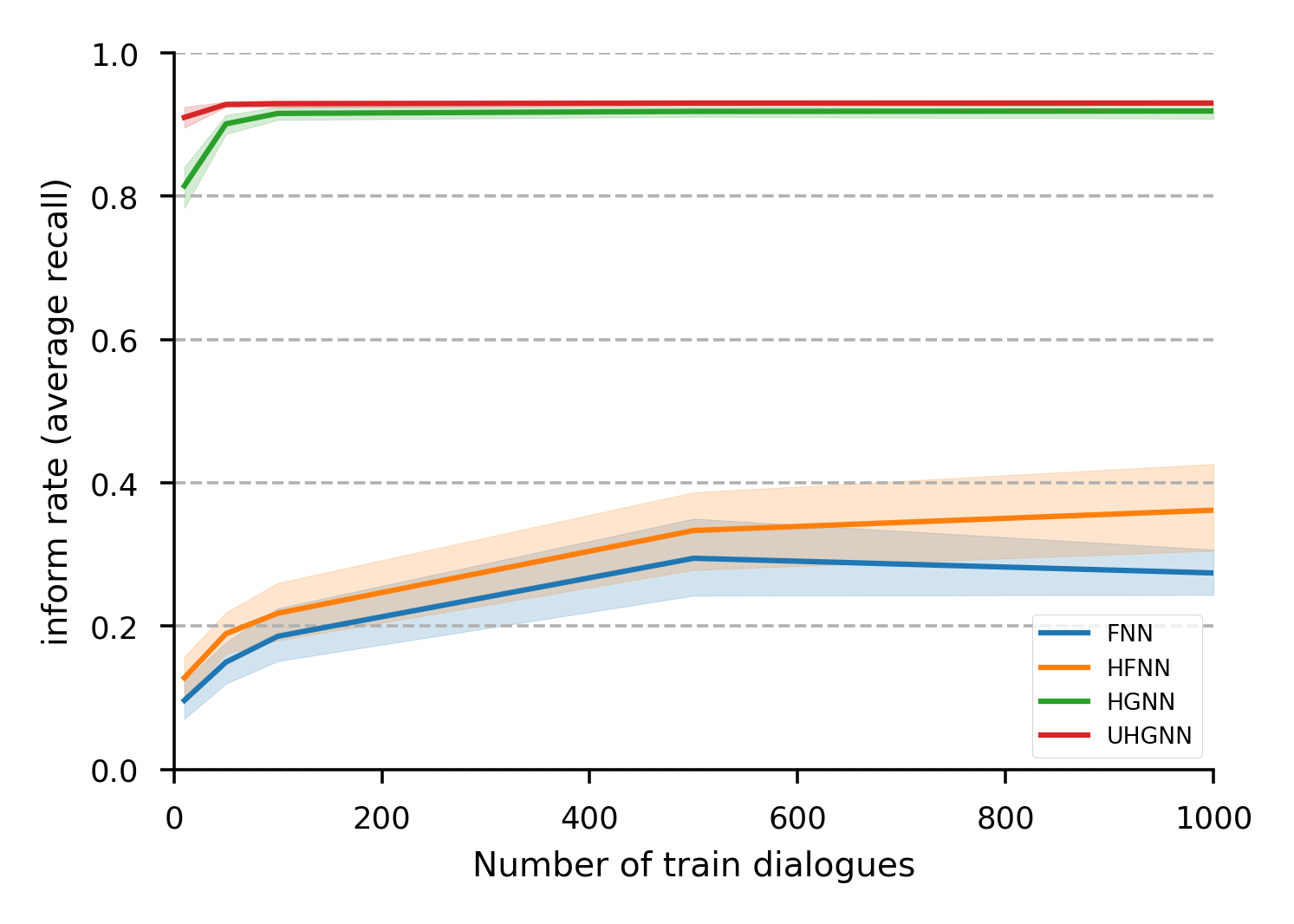}
        \label{subfig:find_CL_pipeline}
        }
        \subfloat[\scriptsize{Book rate.}]{
        \includegraphics[width=0.32\textwidth]{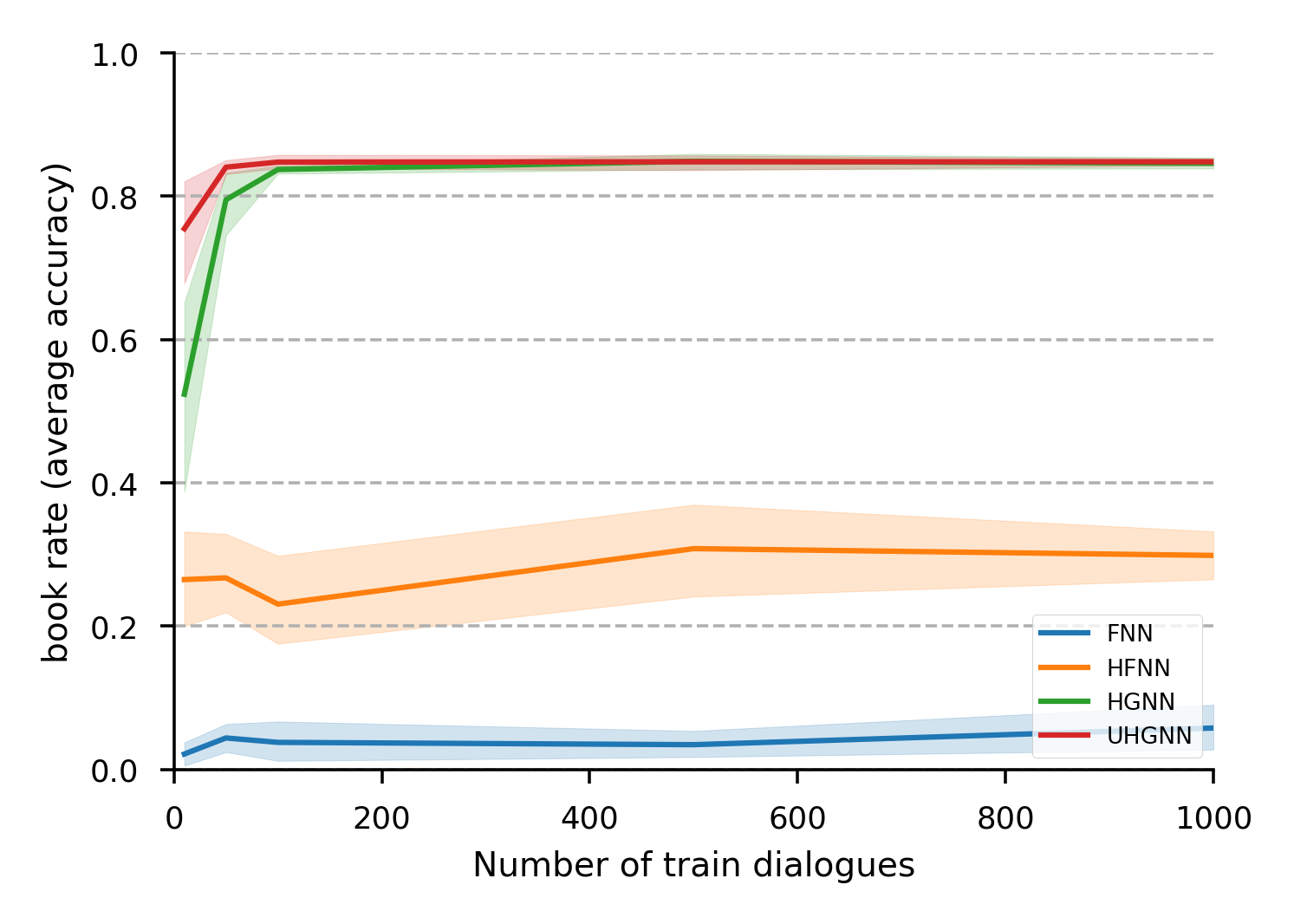}
        \label{subfig:book_CL_pipeline}
        }\\
        \subfloat[\scriptsize{Success rate.}]{
        \includegraphics[width=0.32\textwidth]{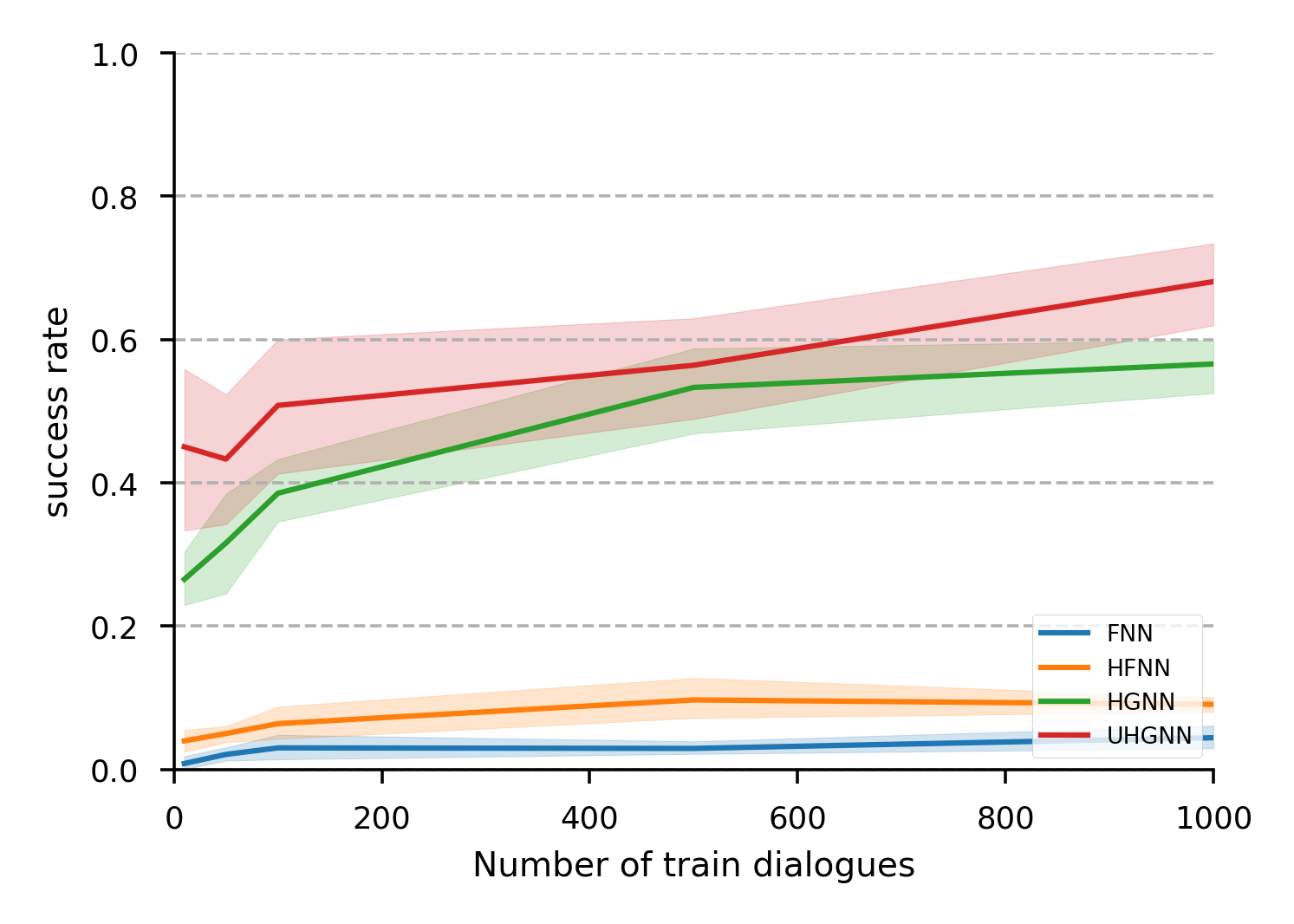}
        \label{subfig:success_MW_pipeline}
        }
        \subfloat[\scriptsize{Inform rate (recall).}]{
        \includegraphics[width=0.32\textwidth]{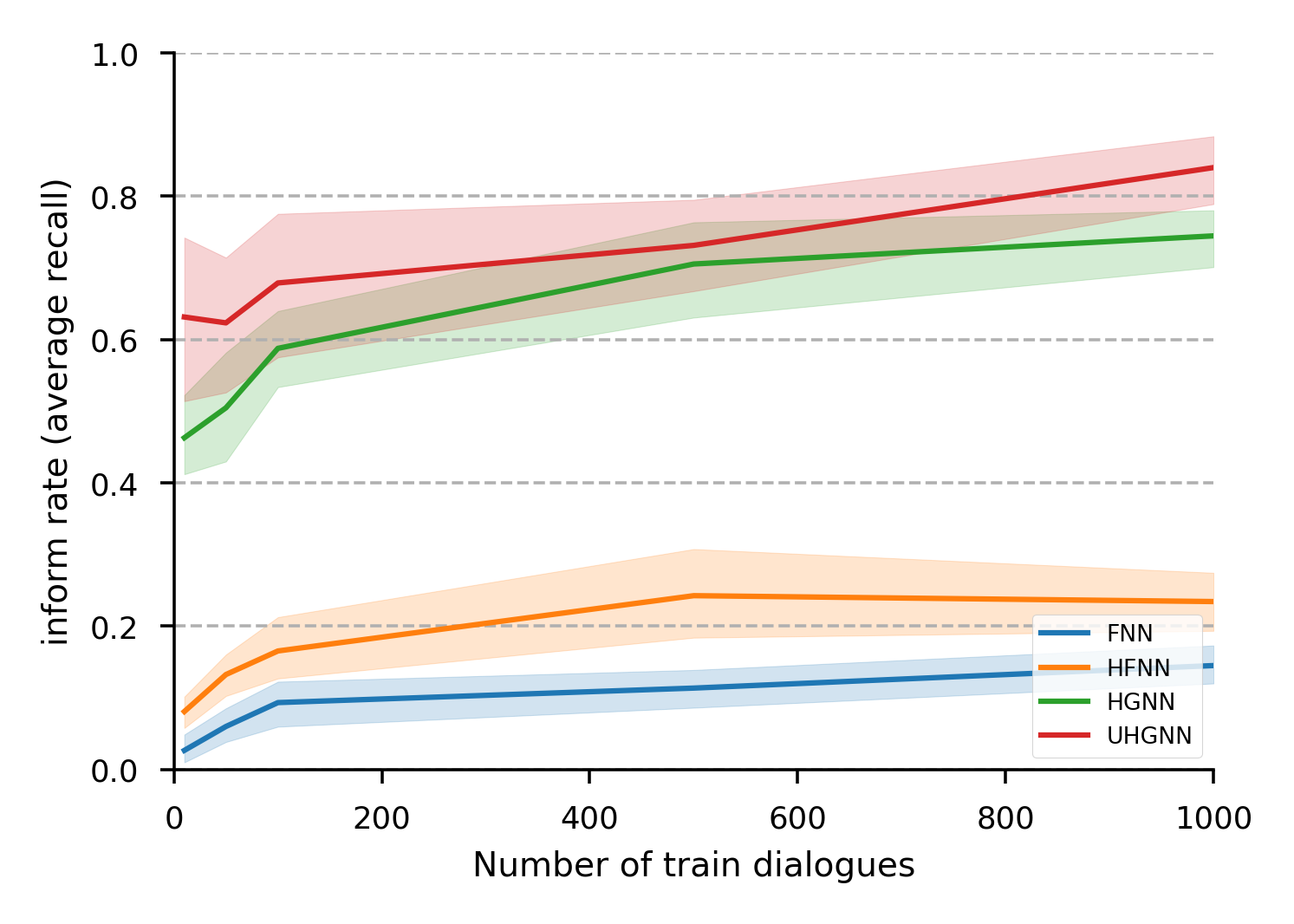}
        \label{subfig:find_MW_pipeline}
        }
        \subfloat[\scriptsize{Book rate.}]{
        \includegraphics[width=0.32\textwidth]{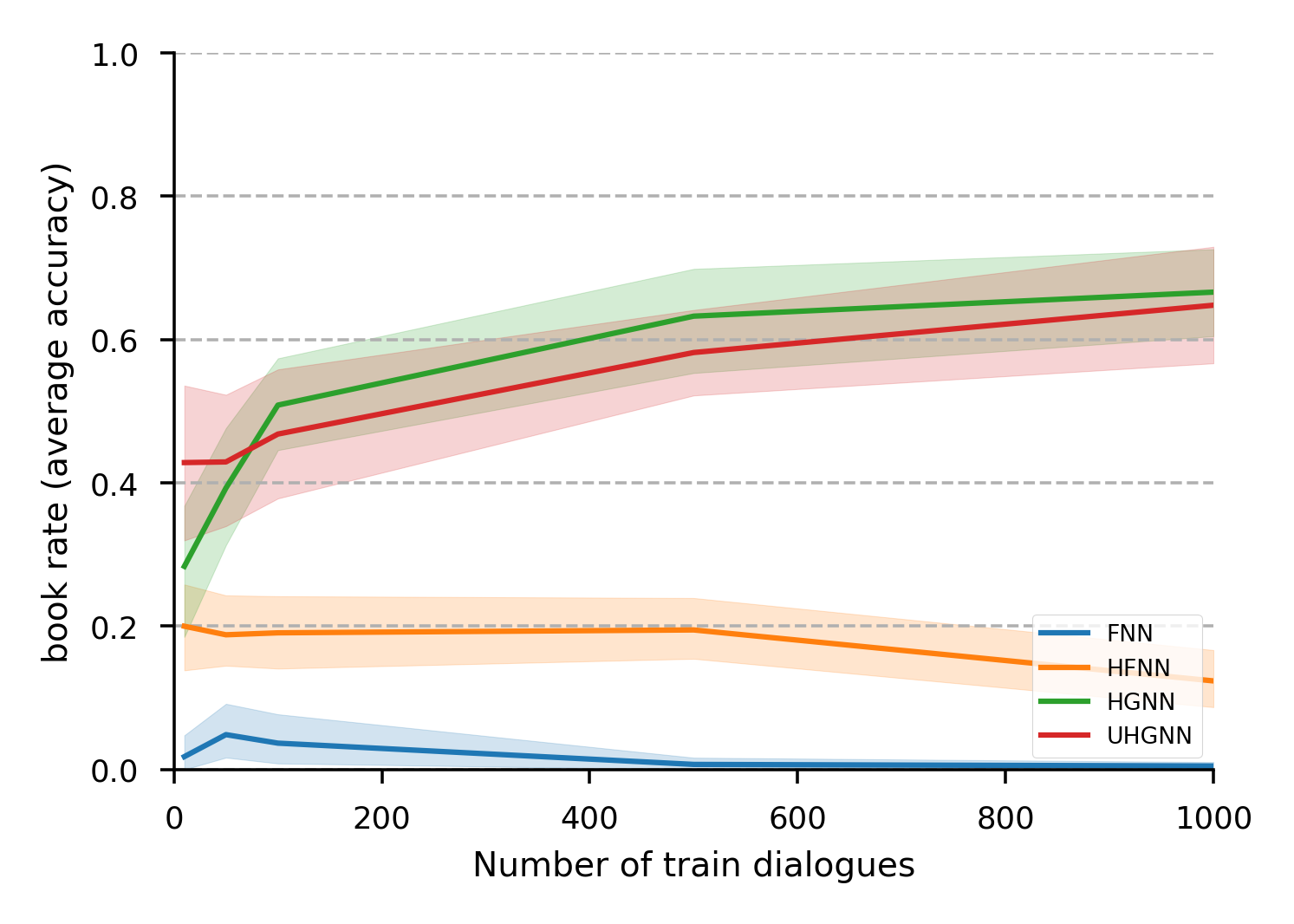}
        \label{subfig:book_MW_pipeline}
        }
    \end{center}
    \caption{Dialogue system performance with simulated user based on $10\,000$ training iterations as a function of the number of training dialogues in a long training scenario. The supervised DM is based on simulated demonstrations (Figures (a),(b),(c)) or on human demonstrations (Figures (d),(e),(f)). The line plot represents the mean and the coloured area represents the $95\%$ confidence interval over a sample of $10$ runs.}
    \label{fig:performance_system}
\end{figure*}

\paragraph{Few-Shot}

We extended the ablation study in a few-shot scenario focusing on the ability of the models to succeed on specific dialogue tasks as reported in Figure~\ref{fig:performance_long}.
In particular, we show the success rate in Figure~\ref{subfig:success_CL_long}, the inform rate (recall) in Figure~\ref{subfig:find_CL_long} and the book rate in Figure~\ref{subfig:book_CL_long} when using simulated demonstrations
and respectively in Figure~\ref{subfig:success_MW_long}, Figure~\ref{subfig:find_MW_long} and Figure~\ref{subfig:book_MW_long} when using human demonstrations.
The more structured the model, the greater the learning efficiency and the greater the data efficiency.
Likewise, we notice that learning is more data-intensive when imitating human strategies. It appears that the booking task is more difficult to perform according to human demonstrations (when comparing Figure~\ref{subfig:book_CL_long} and Figure~\ref{subfig:book_MW_long}) or using a flat architecture (\textsc{FNN} gets null results). We therefore foresee that more high quality data is needed to learn on human dialogues.

\subsection{Dialogue System Evaluation}
\label{subsec:dialogue_system_evaluation}

We continue our analysis on the robustness of the studied models with the entire dialogue system facing both simulated and human users. The dialogue system utilises a \textsc{BERT} \textsc{NLU} \cite{devlin2018bert} and a hand-crafted \textsc{NLG}.

\paragraph{Simulated User Evaluation}

As in the previous subsection, we study the robustness of the models in a few-shot scenario as presented in Figure~\ref{fig:performance_system}. We observe that \textsc{FNN} (in blue) and \textsc{HFNN} (in orange) learning is collapsing when using simulated dialogues (see Figures~\ref{subfig:success_CL_pipeline}, \ref{subfig:find_CL_pipeline} and \ref{subfig:book_CL_pipeline}). On the opposite, \textsc{HGNN} (in green) and \textsc{UHGNN} (in red) performance appears more stable in the entire dialogue system even when using real dialogues (see Figures~\ref{subfig:success_MW_pipeline}, \ref{subfig:find_MW_pipeline} and \ref{subfig:book_MW_pipeline}). 
Therefore, these results confirm that behaviour cloning is easier from simulated than human experts. As observed before in Subsection~\ref{subsec:dialogue_system_evaluation}, this can be explained by an large variability of human strategies (hence the need for more data to improve performance). Another explanation is that simulated dialogues are more in line with the artificial evaluator provided in the \textsc{ConvLab}. In addition, it is important not to neglect the side effects of cascading errors due to successive \textsc{NLU}, \textsc{DST}, \textsc{DM} and \textsc{NLG} modules. In particular, the \textsc{NLU BERT} proposed by \textsc{ConvLab} was pre-trained and evaluated on $7\,372$ user utterances with $14\%$ of errors (F1 $86.4\%$, precision $85.1\%$, recall $87.8\%$). This problem can therefore be exacerbated by cascading human errors, as confirmed in the next paragraph.

Finally, we present a detailed comparison table with the best structured policies \textsc{UHGNN} trained on simulated dialogues of \textsc{ConvLab} noted \textsc{MLE-UHGNN-HDC} (\textsc{HDC} for \textit{hand-crafted policy}) and trained on real dialogues of \textsc{MultiWOZ} noted \textsc{MLE-UHGNN-MW} and the baselines of \textsc{ConvLab} (see Table \ref{tab:sotatableconvlab}). In particular, the \textit{maximum likelihood estimator} (\textsc{MLE}) proposed by \textsc{ConvLab} is an implementation of \textsc{FNN} model trained on \textsc{MultiWOZ} corpus in a very long training scenario (multiple passes on all $10k$ dialogues)\footnote{Another difference is that our models returns one unique action per turn instead of a group of actions.}.
Our models show competitive results against \textsc{ConvLab}'s baselines, confirming that the structured with supervised learning in few-shot settings is adapted to address the difficulties in multi-task multi-domain dialogues.

\begin{table*}[!htb]
\centering
\resizebox{1.95\columnwidth}{!}{%
\begin{tabular}{c|ccccccccc}
    \toprule
    \multirow{2}{*}{\textbf{Configuration}} & \textbf{Avg Turn} & \textbf{Inform rate (\%)} & \textbf{Book} & \multicolumn{2}{c}{\textbf{Complete}} & \multicolumn{2}{c}{\textbf{Success}} \\
    & \textbf{(succ/all)} & \textbf{Prec. / Rec. / F1} & \textbf{Rate (\%)} & \multicolumn{2}{c}{\textbf{Rate (\%)}} & \multicolumn{2}{c}{\textbf{Rate (\%)}} \\
    \midrule \midrule
    \multicolumn{8}{c}{\textbf{Dialogue Management}}\\
    \midrule
    HDC & 10.6/10.6 & 87.2 / 98.6 / 90.9 & 98.6 & \textbf{97.9} & - & \textbf{97.3} & - \\
    MLE-UHGNN-HDC (ours) & 12.8/13.0 & 95.3 / 98.8 / 96.4 & 98.5 & 97.3 & (-0.6) & 95.4 & (-1.9) \\
    MLE-UHGNN-MW (ours) & 16.5/20.7 & 94.3 / 90.7 / 91.6 & 76.7 & 81.4 & (-16.5) & 81.0 & (-6.3) \\
    \midrule
    \multicolumn{8}{c}{\textbf{Dialogue System (BERT NLU + hand-crafted NLG)}} \\
    \midrule
    HDC & 11.4/12.0 & 82.8 / 94.1 / 86.2 & 91.5 & 92.7 & - & 83.8 & - \\
    HDC$^{\dagger}$ & 11.6/12.3 & 79.7 / 92.6 / 83.5 & 91.1 & 90.5 & (-2.2) & 81.3 & (-2.5) \\
    \textsc{MLE}$^{\dagger}$ & 12.1/24.1 & 62.8 / 69.8 / 62.9 & 17.6 & 42.7 & (-50.0) & 35.9 & (-47.9) \\
    \textsc{PG}$^{\dagger}$ & 11.0/25.3 & 57.4 / 63.7 / 56.9 & 17.4 & 37.4 & (-55.3) & 31.7 & (-52.1) \\
    \textsc{GDPL}$^{\dagger}$ & 11.5/21.3 & 64.5 / 73.8 / 65.6 & 20.1 & 49.4 & (-43.3) & 38.4 & (-45.4) \\
    \textsc{PPO}$^{\dagger}$ & 13.1/17.8 & 69.4 / 85.8 / 74.1 & 86.6 & 75.5 & (-17.2) & 71.7 & (-12.1) \\
    \textsc{MLE-UHGNN-HDC} (ours) & 14.0/15.4 & 89.3 / 93.0 / 90.2 & 84.8 & \textbf{90.0} & (-2.7) & \textbf{82.7} & (-1.1) \\
    \textsc{MLE-UHGNN-MW} (ours) & 17.0/23.0 & 84.0 / 87.6 / 84.5 & 64.8 & 72.1 & (-20.6) & 68.1 & (-15.7) \\
    \bottomrule
\end{tabular}
}
\caption{Dialogue manager and system evaluations with simulated users. When evaluating the dialogue manager, the simulated user passes directly dialogue acts and vice-versa. Our tested configurations
are evaluated and averaged on $10$ run each with $250$ dialogues. Configurations with ${\dagger}$ are taken from the \href{https://github.com/thu-coai/ConvLab-2/tree/ad32b76022fa29cbc2f24cbefbb855b60492985e}{GitHub of \textsc{ConvLab}}.
}
\label{tab:sotatableconvlab}
\end{table*}

\paragraph{Human Evaluation}

\begin{table}[!tb]
\centering
\resizebox{0.95\columnwidth}{!}{%
\begin{tabular}{c|ccc}
    \toprule
    \textbf{Dialogue System} & \textbf{Avg} & \textbf{Satisfaction} & \textbf{Nb of} \\
    \textbf{(BERT NLU + Rule NLG)} & \textbf{Turn} & \textbf{Rate (\%)} & \textbf{Dial.} \\
    \midrule \midrule
    HDC & 22.6 & \textbf{92.6 $\pm$ 9.87} & 27 \\
    MLE-UHGNN-HDC & 25.6 & 50.0 $\pm$ 14.8 & 44 \\
    MLE-UHGNN-MW & 17.3 & 36.7 $\pm$ 17.2 & 30 \\
    \bottomrule
\end{tabular}
}
\caption{Dialogue system evaluation with real users with a 95\% confidence level for satisfaction rate.}
\label{tab:humantableconvlab}
\end{table}

We organised preliminary evaluation sessions, in which volunteers were invited to chat on-line with three dialogue systems that were randomly assigned\footnote{Crowdsourcing was not used because of ethical concerns regarding the work conditions of collaborators. Volunteers from our research institution were invited to participate and they were aware of the scientific motivations behind the evaluation. In this sense, they were motivated to participate without any economic reward implying no pressure and without knowing the nature of the models they were evaluating, avoiding in this way any evaluation bias.}. Subjects do not know which system they are evaluating. Each system has a different \textsc{DM} model: \textsc{HDC} (\textit{hand-crafted policy}), \textsc{MLE-UHGNN-HDC} (based on simulated demonstrations with \textsc{HDC} policy) and \textsc{MLE-UHGNN-MW} (based on \textsc{MultiWOZ} demonstrations) combined with the \textsc{BERT NLU} and the hand-crafted \textsc{NLG} provided by \textsc{ConvLab}. At the end of the chat, evaluators were asked whether or not they reach the goal and were satisfied with the performance of the system. The \textbf{satisfaction rate} is then the proportion of dialogues in which the system solved the task at the end of the dialogue according to the human evaluator. We reported results on roughly $30$ dialogues for each method. The results of this experimentation are presented in Table~\ref{tab:humantableconvlab}. Although test is small-sized and not highly statistically significant, these preliminary results are disconcerting with respect to the simulated ones. The \textsc{HDC} does very well whereas \textsc{MLE-UHGNN-HDC} gets by in half the cases, \textsc{MLE-UHGNN-MW} fails in most cases.

These results can be explained by the limitations of the \textsc{NLU} facing impatient evaluators, short and ambiguous sentences where the active domain is unclear (as in this example of the user saying "What is the name?") or typographical errors.
Moreover, it is important to underline that \textsc{ConvLab} does not natively propose the management of uncertainties in the state representation which can strongly restrict the performance of the learning methods in noisy environments.
Another limitation is that the \textsc{HDC} is more adapted to
conventional dialogues whereas \textsc{MLE-UHGNN}s were trained only on winning dialogues. This implies that learning methods are more sensitive to dialogues that break out of the learned patterns.
Similarly, the strategies of simulated and real users do not seem to be well aligned with each other and even more strongly with the expectations of human evaluators.

\section{Conclusion}
\label{sec:conclusion}

We investigated in this work the impact of policy structure and experts on success rate in {few-shot learning} for multi-domain multi-task dialogues.
Promising results were obtained: hierarchical and generic \textsc{GNN} policies are able to achieve remarkable performance with few dialogues and few training iterations when following a simulated expert. This confirms the growing interest for these neural structures.
We also present an important finding: the policy performance degrades in few-shot learning when using human demonstrations. This fact questions the alignment between dialogue evaluators and human strategies in state-of-the-art dialogue frameworks.

\section*{Limitations}

The reduced performance when learning from human experts suggests that we shall concentrate the efforts in bridging the gap between automatic evaluators and high-quality human-human datasets.  We also devise the use of \textit{curriculum learning} \cite{bengio2009curriculum} strategies: starting from simple -- simulated -- dialogues then adding progressively more complex, human dialogues demonstrations.

It is also necessary to analyse the impact of \textsc{GNN} policies with neural \textsc{NLU}/\textsc{NLG} modules to study how to integrate such structures in end-to-end architectures.

We point out some limitations of \textsc{ConvLab}. The detection of the active domain is sensitive to the output of the \textsc{NLU} and thus sensitive to ambiguous statements. Data representation restricts the \textsc{DST} to a deterministic view and must be adapted to a probabilistic representation to capture the uncertainties in the user's input. Similarly, it may be worthwhile to improve the action space by adding more possibilities for human users, for instance to \textsc{confirm} or \textsc{deny} in a more flexible way.

Finally, the human evaluation was performed on a small scale and on models trained in a context with few training iterations. A more in-depth or supervised study could shed more light on the raised issues.





\section*{Acknowledgements}
This work was performed using HPC resources from GENCI-IDRIS (Grant 20XX-[AD011011407]).
\bibliography{anthology,custom}
\bibliographystyle{acl_natbib}

\appendix

\end{document}